\documentclass[onecolumn, 11pt]{protocol}
\usepackage{blindtext}

\leadauthor{Müller}

\usepackage[framemethod=TikZ]{mdframed}
\usepackage{listings}
\usepackage{wrapfig}
\usepackage{booktabs}
\usepackage{tcolorbox}
\usepackage{array}
\usepackage{enumitem}

\newmdtheoremenv [ %
outerlinewidth=1,%
roundcorner=2pt,%
leftmargin=40,%
rightmargin=40,%
backgroundcolor=blue!10,%
outerlinecolor=blue!40,%
innertopmargin=\topskip,%
splittopskip=\topskip,%
]{coloredbox}{Box}

\newenvironment{protobox}[2][t]
{\begin{table}[#1]\begin{coloredbox} \textbf{#2}\\}  
{\end{coloredbox}\end{table}}

\newenvironment{demodata}[1]
{\begin{tcolorbox}[boxrule=1pt,colback=green!10,arc=2pt]
\textsc{Demo Data:} #1\\
}  
{\end{tcolorbox}}

\AtEveryBibitem{\clearlist{language}\clearlist{note}\clearlist{editor}}

\newcommand{\sizetwo}[2]{\ensuremath{#1\!\times\!#2\xspace}}
\newcommand{\sizethree}[3]{\ensuremath{#1\mkern-2mu\times\mkern-2mu#2\mkern-2mu\times\mkern-2mu#3\xspace}}

\makeatletter
\DeclareRobustCommand\onedot{\futurelet\@let@token\@onedot}
\def\@onedot{\ifx\@let@token.\else.\null\fi\xspace}

\def\eg{\emph{e.g}\onedot} \def\Eg{\emph{E.g}\onedot}
\def\ie{\emph{i.e}\onedot} 
\def\cf{\emph{cf.}\xspace}

\makeatother

\newcommand{\um}{\ensuremath{\mu m}\xspace}

\newcommand{\figOverview}{
\begin{figure}[t]
\centering
\includegraphics[width=1\linewidth]{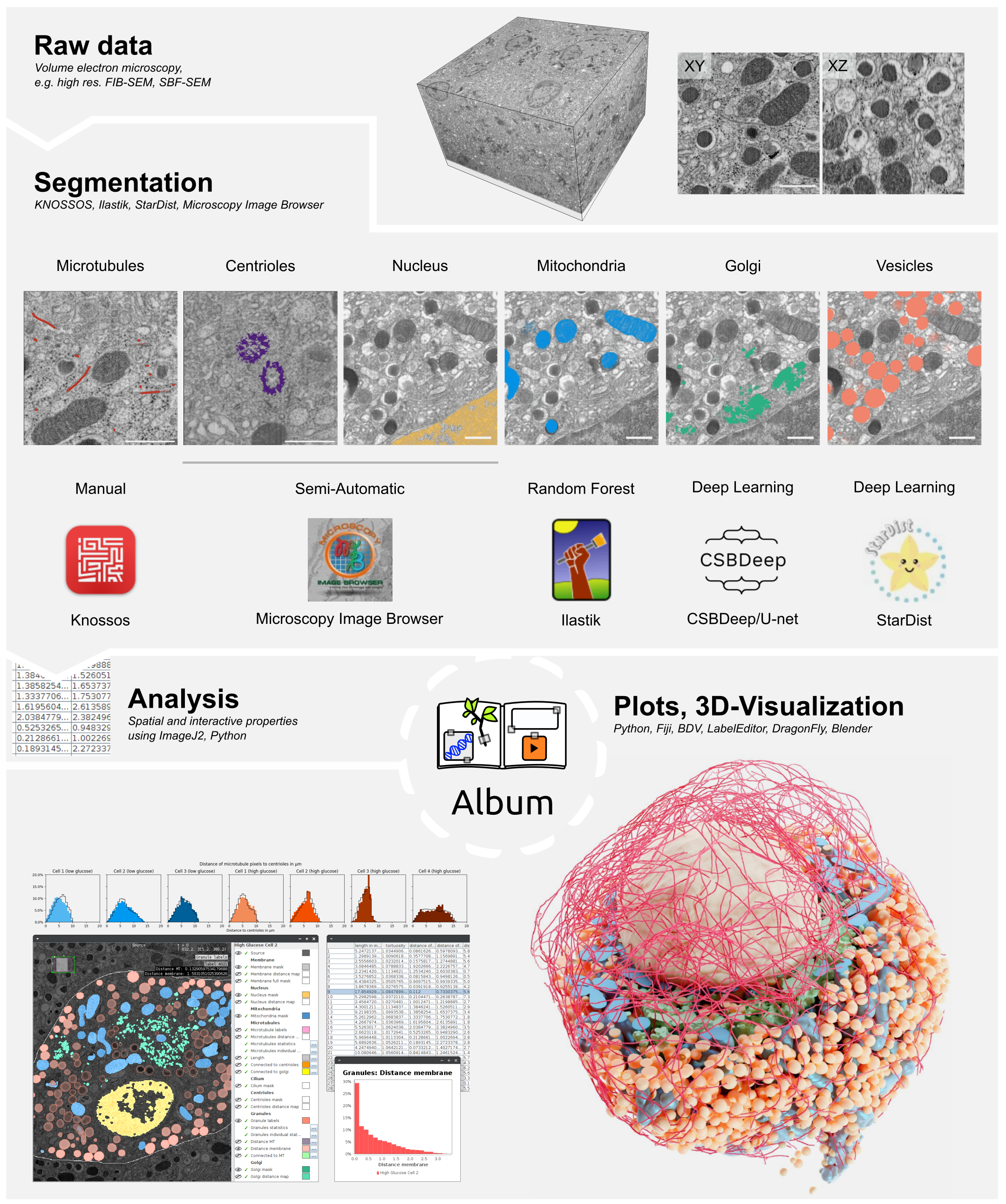}
\caption{\textbf{Overview of the protocol.} Workflow starting with raw vEM data followed by organelle-specific segmentation, spatial analysis of the data and 3D rendering of the results. Analysis, plotting and 3D rendering steps can be performed via Album. Raw data and segmentation panel adapted from~\cite{muller2021} under CC BY 4.0 license.}
\label{fig:overview}
\end{figure}
}

\newcommand{\figSegmentation}{
\begin{figure}[t]
\centering
\includegraphics[width=1\linewidth]{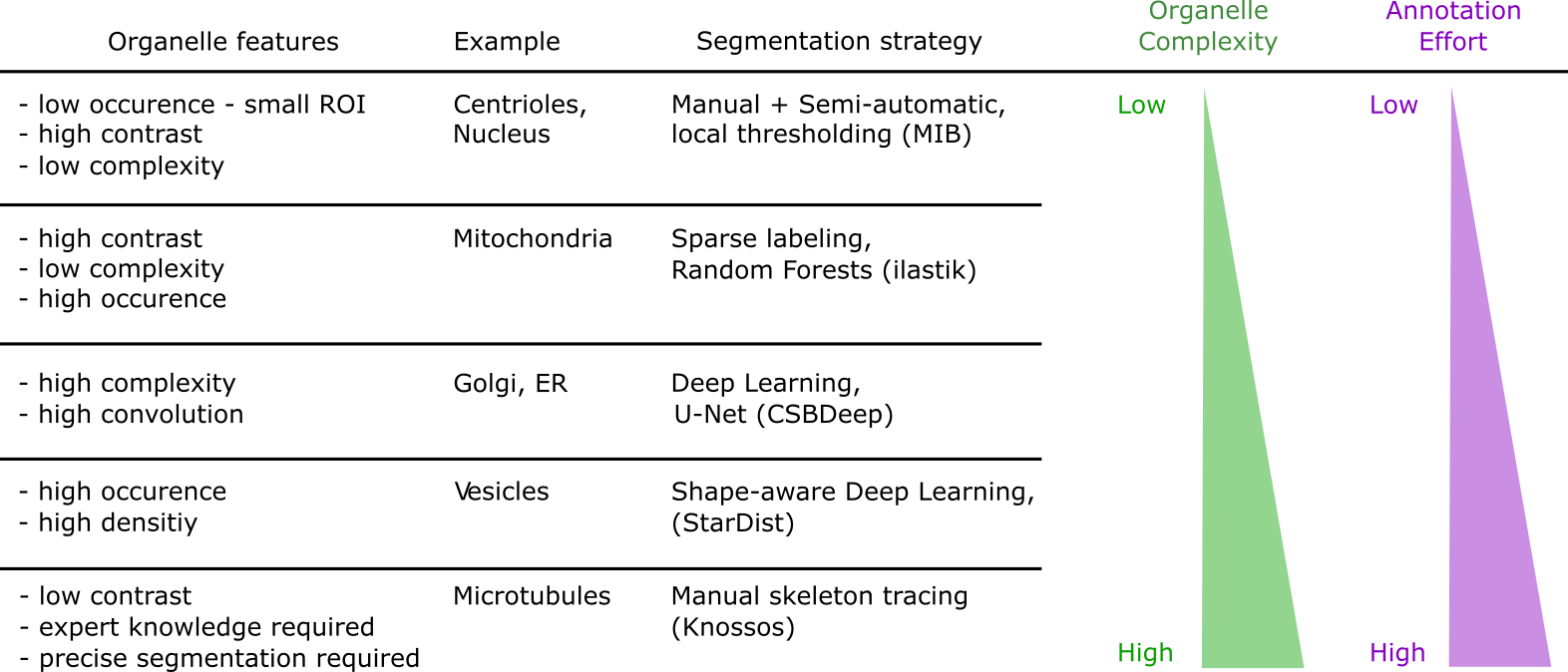}
\caption{\textbf{Guidelines for choosing segmentation approaches.}
Different organelles have characteristic features that should be taken into account when deciding for a segmentation approach. This enables the choice of time-and workload-efficient strategies.
}
\label{fig:Segmentation}
\end{figure}
}

\newcommand{\figRaw}{
\begin{figure}[t]
\centering
\includegraphics[width=1\linewidth]{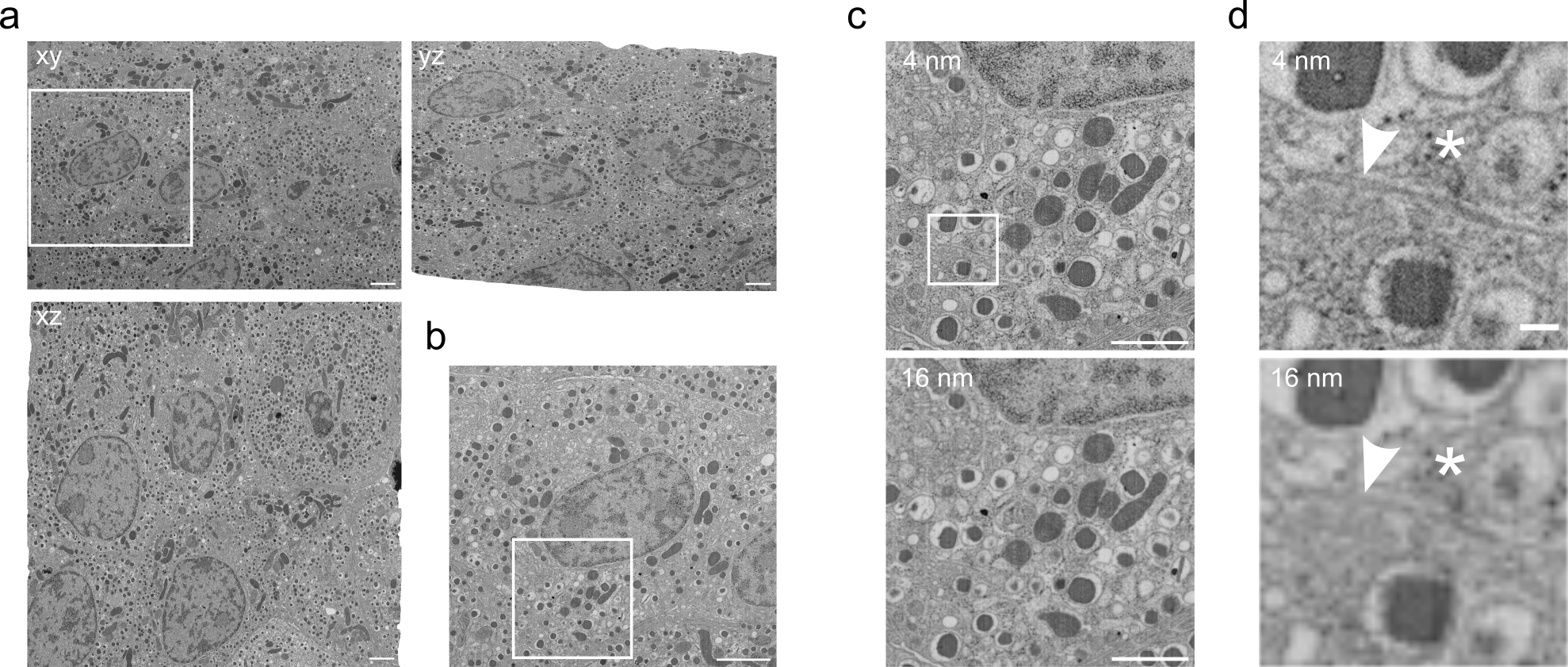}
\caption{\textbf{Preparation of raw data for segmentation.}
a) Full FIB-SEM volume of pancreatic islets containing several cells in xy, xz and yz view. Scale bar: 2 µm.
b) The cell in the marked region of a) is cut out in \Fiji. Scale bar: 2 µm
c) Magnified view of boxed region of b at the original resolution (4 nm voxels) and after binning (16 nm voxels). Scale bar: 1 µm
d) Further magnifications to demonstrate the differences in resolution. Microtubules (arrowhead) and ribosomes (asterisk) are barely visible at 16 nm voxels. Scale bar: 100 nm.
}
\label{fig:raw}
\end{figure}
}

\newcommand{\figMIB}{
\begin{figure}[t]
\centering
\includegraphics[width=1\linewidth]{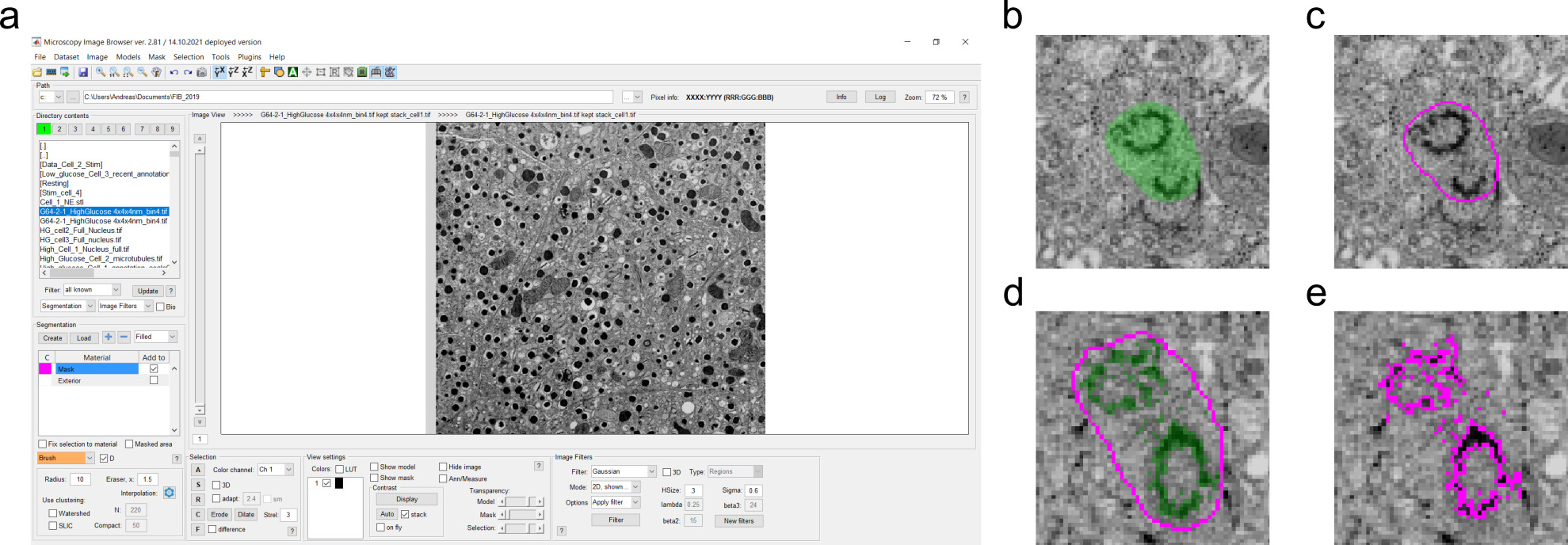}
\caption{\textbf{Local thresholding with MIB.}
a) Overview of the MIB window with raw data.
b) XY view of the data with rough annotation surrounding the centrioles.
c) Area of b after converting the annotation into a mask (magenta).
d) Local thresholding (green) within the masked area.
e) Conversion of of the area to a mask which can be exported.}
\label{fig:MIB}
\end{figure}
}

\newcommand{\figKnossos}{
\begin{figure}[t]
\centering
\includegraphics[width=1\linewidth]{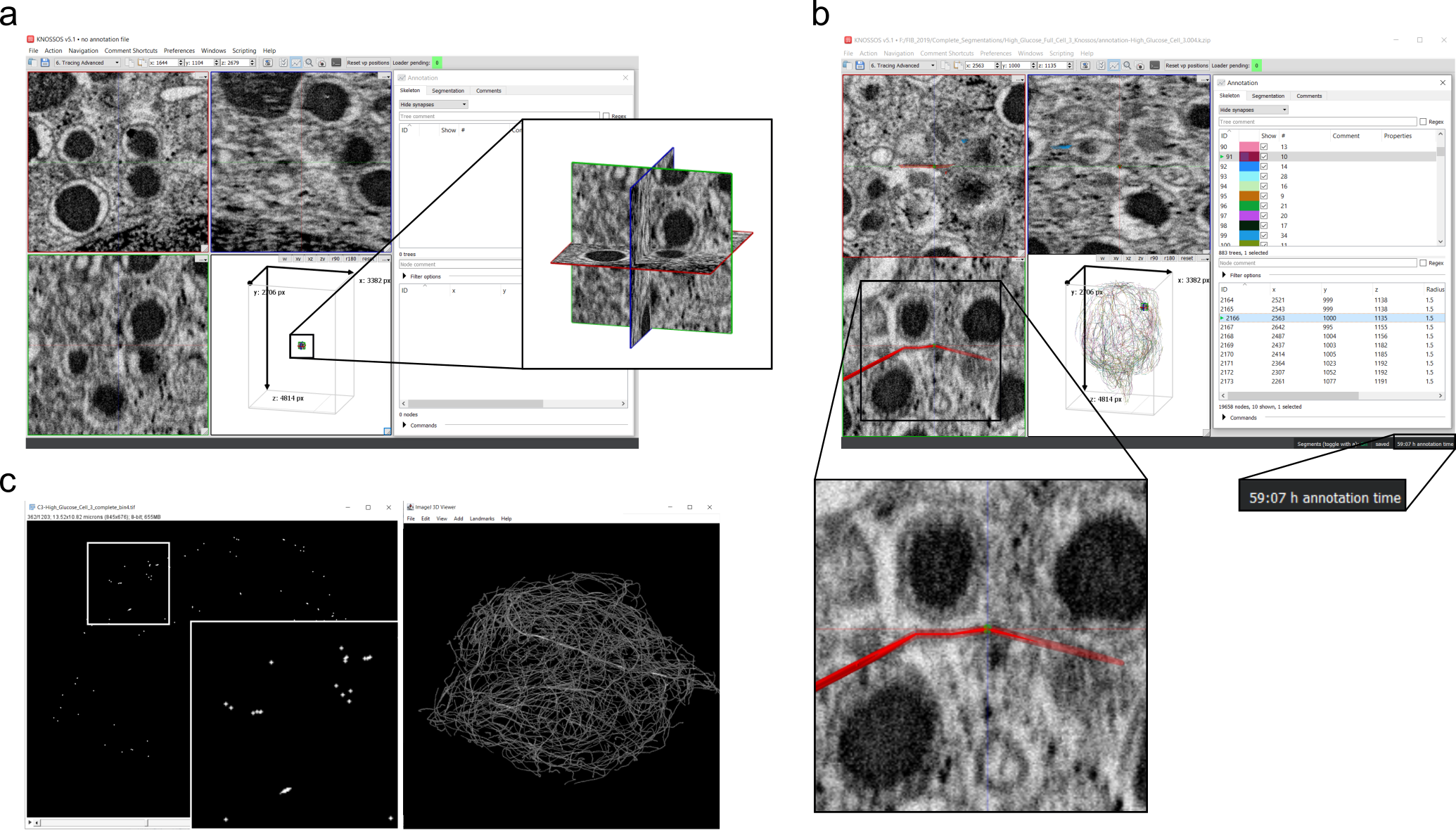}
\caption{\textbf{Skeleton Tracing with \knossos.}
a) Overview of the \knossos window with raw data. Inset highlights the portion of the data that is currently loaded into memory.
b) \knossos window including final skeleton trace of the cell. On the left you see the annotation window with single traces in different colors and below the nodes of the active trace. Insets highlight one trace with a node and the annotation time.
c) Skeleton trace after conversion to binary mask in \Fiji with single slice through the stack on the left and 3D visualization in the 3D viewer on the right.}
\label{fig:knossos}
\end{figure}
}

\newcommand{\figIlastik}{
\begin{figure}[t]
\centering
\includegraphics[width=1\linewidth]{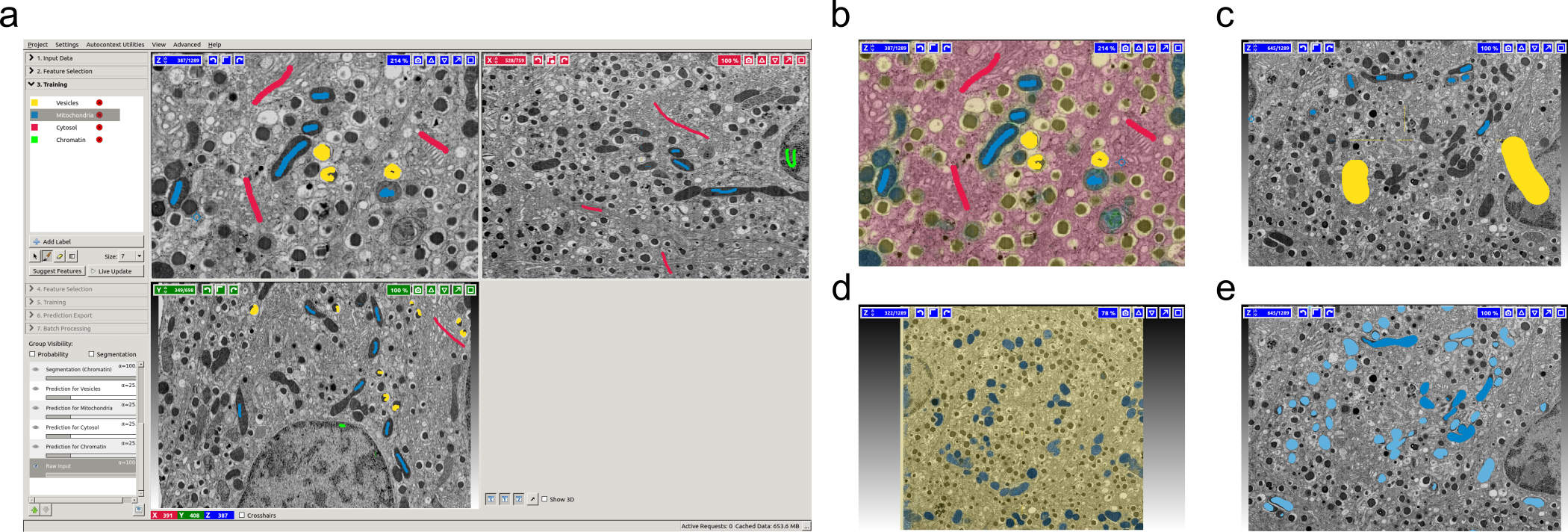}
\caption{\textbf{Autocontext segmentation in ilastik.}
a) Overview of the ilastik window with raw data and sparse
annotations for 4 distinct structures. b) XY view of the data after the first round of training with the different
probability in the color-code according to the labeling. c) XY view for the second round of annotation with
labels for background (yellow) and mitochondria (blue). d) XY view after the second round of training with
probabilities for background (yellow) and mitochondria (blue). e) Object detection of mitochondria (blue).}
\label{fig:ilastik}
\end{figure}
}

\newcommand{\figDL}{
\begin{figure}[t]
\includegraphics[width=1\linewidth]{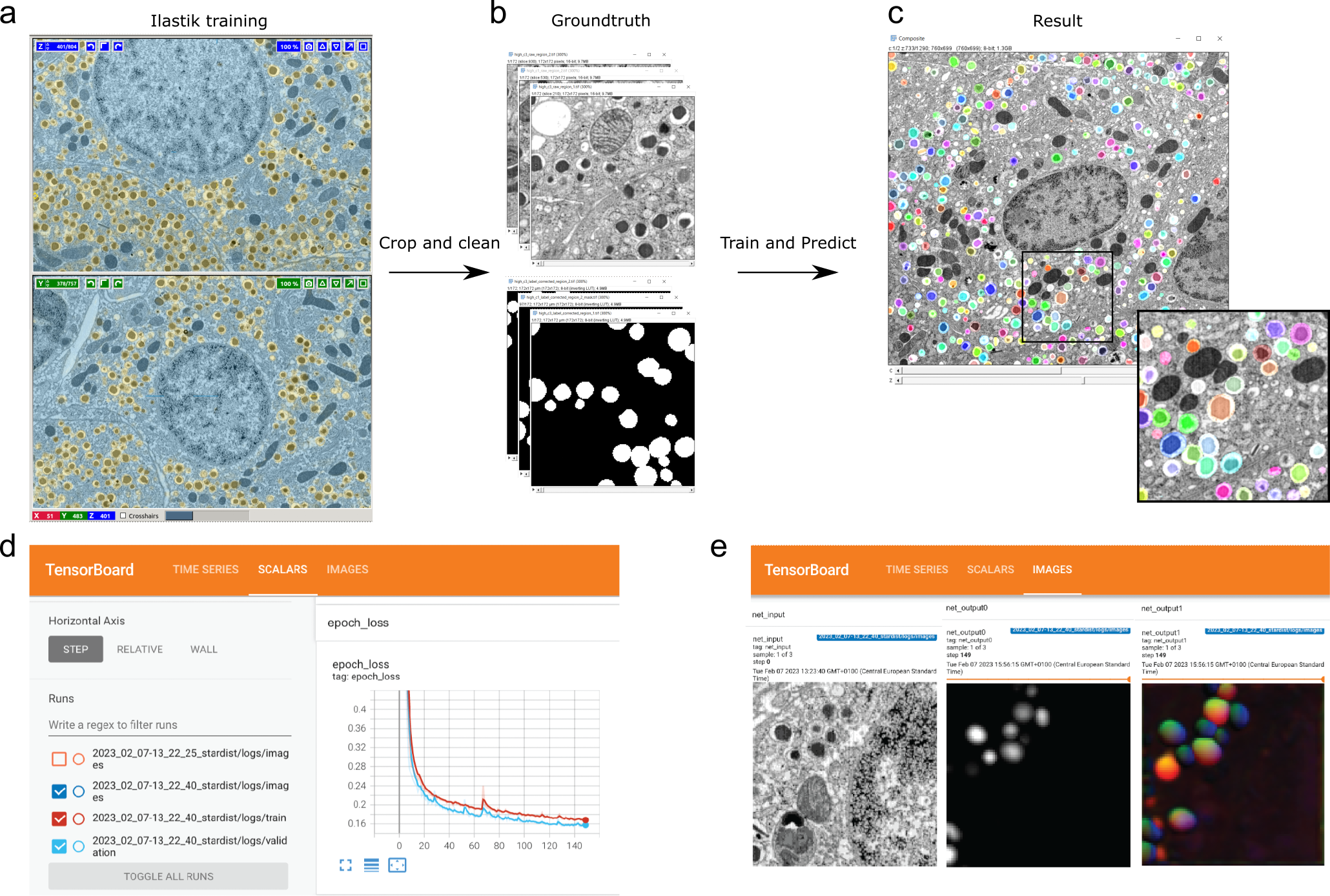}
\caption{\textbf{Segmentation of crowded insulin SGs with \stardist.}
a) Initial segmentation of SGs via the autocontext workflow in ilastik yields preliminary ground truth masks. b) Final ground truth masks are  generated by selecting small crops and manually curating the labels. c) After training a \stardist model it ca be used to predict on new, full sized stacks. d) Monitoring the \stardist model training via \texttt{tensorboard}. Shown are the loss curves for both training and validation, which should gradually decrease. e)  \texttt{tensorboard} will as well show an example input slice (left) and the corresponding intermediate model outputs (center: object probability, right: distance maps) which both should get more and more refined during training. } 
\label{fig:DL}
\end{figure}
}

\newcommand{\figrendering}{
\begin{figure}[t]
\centering
\includegraphics[width=1\linewidth]{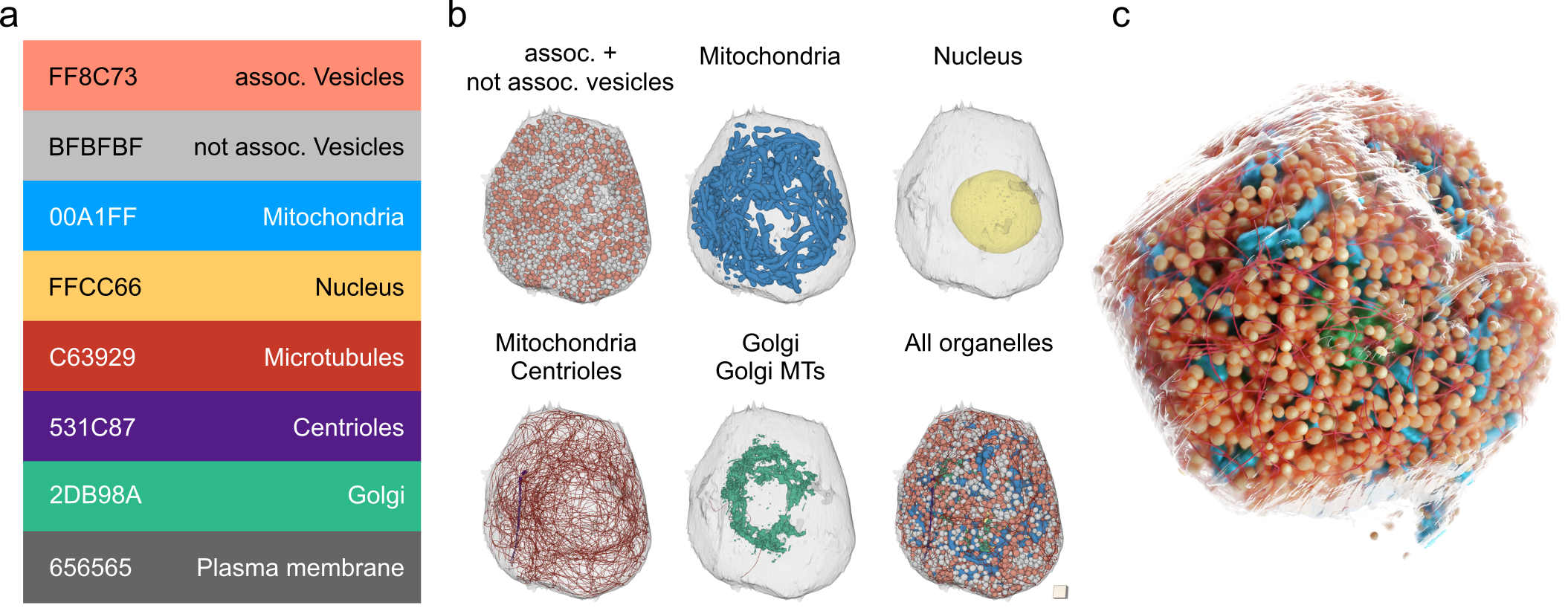}
\caption{\textbf{3D visualisation of vEM data}
a) Color palette with HTML color codes and corresponding organelles.
b) 3D renderings of organelles shown separately and as complete cell in ORS Dragonfly. ({\small Panel adapted from [10] under CC BY 4.0 license.})
c) 3D rendering of the same cell as in b) with blender.}
\label{fig:3D}
\end{figure}
}

\newcommand{\figanalysisimport}{
\begin{figure}[t]
\centering
\includegraphics[width=1\linewidth]{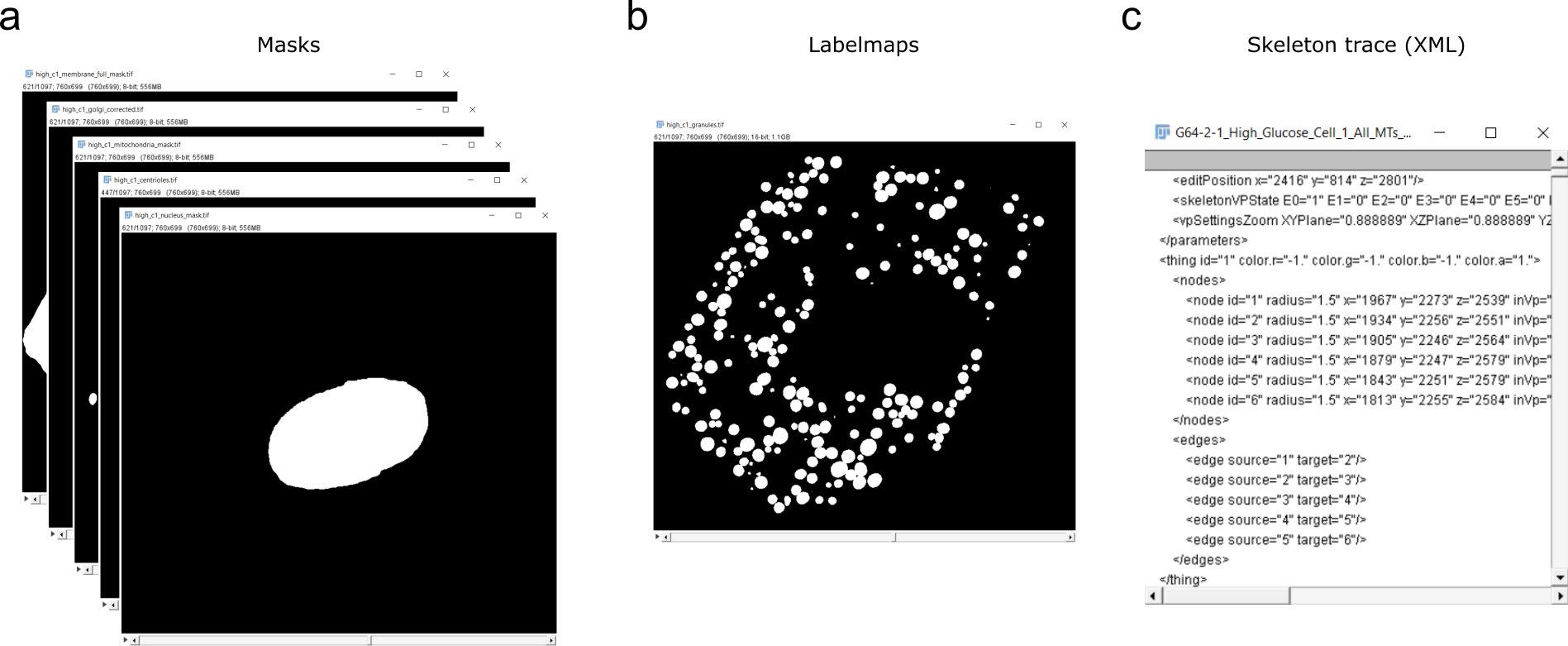}
\caption{\textbf{Types of data to be imported into the analysis project.} a) 8 bit TIFF binary masks. b) 16 bit labelmaps. c) Skeleton traces XML files.}
\label{fig:analysisimport}
\end{figure}
}

\newcommand{\figcellviewer}{
\begin{figure}[t]
\centering
\includegraphics[width=1\linewidth]{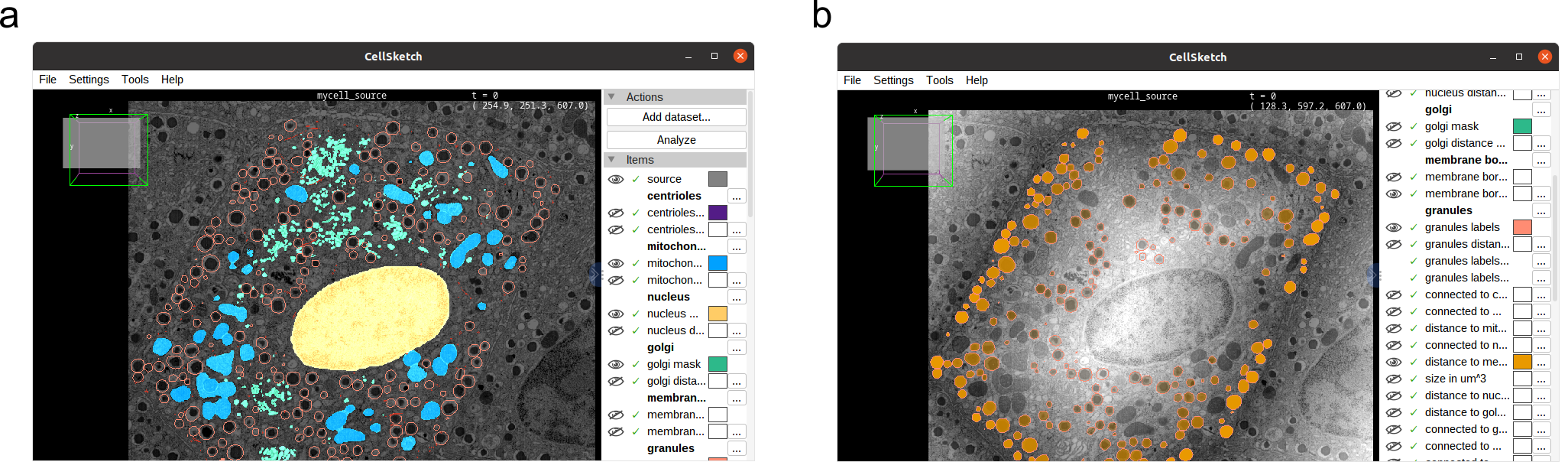}
\caption{\textbf{\cellsketch Viewer}
a) The \cellsketch Viewer displaying the datasets imported into a new \cellsketch project.
b) The \cellsketch Viewer displaying analysis results of a \cellsketch project, in this case the distance of vesicles to the cell membrane and the distance map of the membrane.
}
\label{fig:cellviewer}
\end{figure}
}

\newcommand{\figanalysisplots}{
\begin{figure}[t]
\centering
\includegraphics[width=1\linewidth]{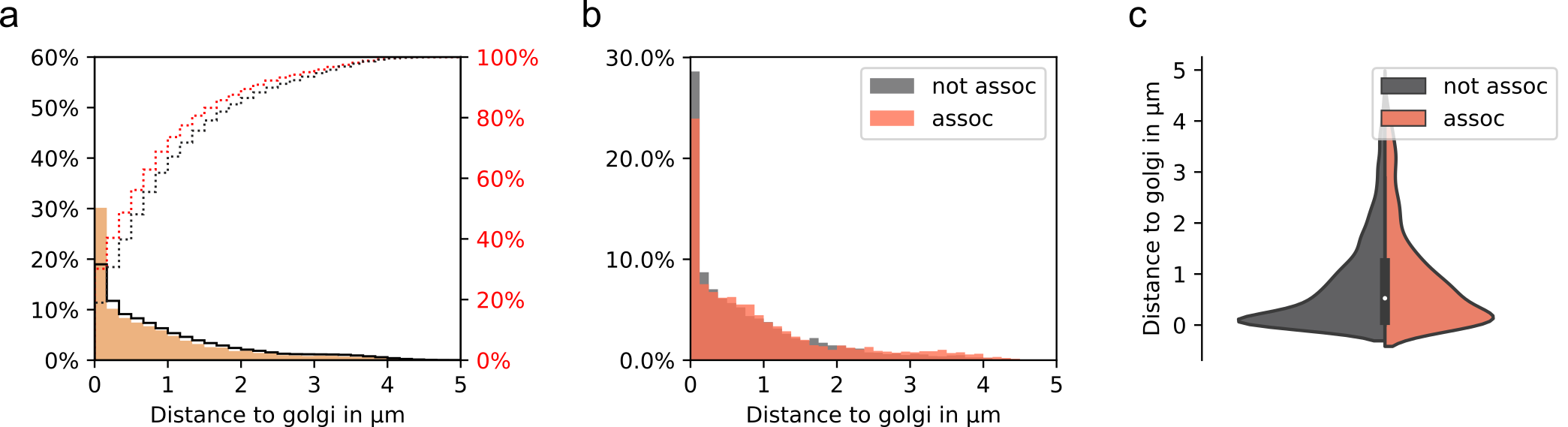}
\caption{\textbf{Plots generated from the spatial analysis results.} a) Distribution of the distances of insulin SGs to the Golgi apparatus. Black line indicates a random distribution. Red dotted and black dotted lines represent actual and random cumulative distributions, respectively. b) Distribution of the distances of microtubule-associated and not-associated insulin SGs to the Golgi apparatus. c) Violin plot showing the distance of  microtubule-associated and not-associated insulin SGs to the Golgi apparatus.}
\label{fig:analysisplots}
\end{figure}
}

\newcommand{\figvtk}{
\begin{figure}[t]
\centering
\includegraphics[width=1\linewidth]{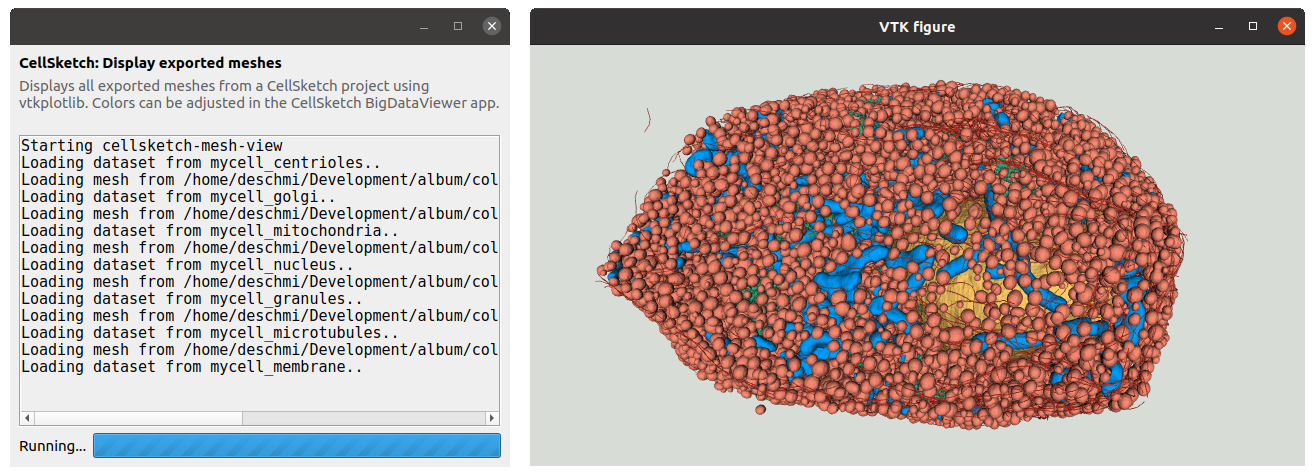}
\caption{\textbf{Exported meshes displayed in VTK}
Cell components from a \cellsketch project are automatically exported as meshes and displayed in VTK with a single solution call.
}
\label{fig:vtk}
\end{figure}
}

\newcommand{\Fiji}{\textsc{Fiji}\xspace}
\newcommand{\stardist}{{\small\textsc{StarDist}}\xspace}
\newcommand{\knossos}{{\small\textsc{Knossos}}\xspace}
\newcommand{\cellsketch}{{\small\textsc{CellSketch}}\xspace}

\newcommand{\menu}{\ensuremath{\triangleright}\xspace}

\newcommand*\circled[3]{\tikz[baseline=(char.base)]{
    \node[shape=circle, fill=#2, draw=#3, line width=1pt, inner sep=1pt] (char) {#1};}}

\newenvironment{steps}[1][]
{\enumerate[label=\protect\circled{\small \arabic*}{orange!10}{orange!60}, leftmargin=4ex,#1]}
{\endenumerate}

\begin{document}

\title{Organelle-specific segmentation, spatial analysis, and visualization of volume electron microscopy datasets}
\shorttitle{Organelle-specific segmentation}

\author[1,2,3,$\dagger$]{Andreas Müller}
\author[4,$\dagger$]{Deborah Schmidt}
\author[4]{Lucas Rieckert}
\author[1,2,3, \Letter]{Michele Solimena}
\author[5, \Letter]{Martin Weigert}

\affil[1]{Molecular Diabetology, University Hospital and Faculty of Medicine Carl Gustav Carus, TU Dresden, Dresden, Germany}
\affil[2]{Paul Langerhans Institute Dresden (PLID) of the Helmholtz Center Munich at the University Hospital Carl Gustav Carus and Faculty of Medicine of the TU Dresden, Dresden, Germany}
\affil[3]{German Center for Diabetes Research (DZD e.V.), Neuherberg, Germany}
\affil[4]{HELMHOLTZ IMAGING, Max Delbrück Center for Molecular Medicine (MDC) in the Helmholtz Association, Berlin, Germany}
\affil[5]{Institute of Bioengineering, School of Life Sciences, École Polytechnique Fédérale de Lausanne (EPFL), Lausanne, Switzerland}
\affil[$\dagger$]{Authors contributed equally.}

\maketitle

\begin{abstract}
Volume electron microscopy is the method of choice for the in-situ interrogation of cellular ultrastructure at the nanometer scale.
Recent technical advances have led to a rapid increase in large raw image datasets that require computational strategies for segmentation and spatial analysis.
In this protocol, we describe a practical and annotation-efficient pipeline  for organelle-specific segmentation, spatial analysis, and visualization of large volume electron microscopy datasets using freely available, user-friendly software tools that can be run on a single standard workstation.
We specifically target researchers in the life sciences with limited computational expertise, who face the following tasks within their volume electron microscopy projects:
\emph{i)} How to generate 3D segmentation labels for different types of cell organelles while minimizing manual annotation efforts,
\emph{ii)} how to analyze the spatial interactions between organelle instances, and  
\emph{iii)} how to best visualize the 3D segmentation results.
To meet these demands we give detailed guidelines for choosing the most efficient segmentation tools for the specific cell organelle.
We furthermore provide easily executable components for spatial analysis and 3D rendering  and bridge compatibility issues between freely available open-source tools, such that others can replicate our full pipeline starting from a raw dataset up to the final plots and rendered images.
We believe that our detailed description can serve as a valuable reference for similar projects requiring special strategies for single- or multiple organelle analysis which can be achieved with computational resources commonly available to single-user setups.

 \end {abstract}


\begin{corrauthor}
michele.solimena@uniklinikum-dresden.de, martin.weigert@epfl.ch
\end{corrauthor}

\section*{Introduction}

For several decades two-dimensional (2D) electron microscopy (EM) has been used to study the intricate details of sub-cellular components, contributing to key findings in cell biology and medicine.
In recent years three-dimensional (3D) volume electron microscopy (vEM) has established itself as a widely adopted new imaging modality, leading to a rapid proliferation of large raw image datasets of cells and tissues~\cite{peddie_exploring_2014, peddie_volume_2022}.
vEM includes methods such as serial section transmission electron microscopy (ssTEM), serial section electron tomography, serial block-face scanning electron microscopy (SBF-SEM), and focused ion beam scanning electron microscopy (FIB-SEM).
The recent popularity of vEM has been made possible by major advances in sample preparation~\cite{hua_large-volume_2015} as well as significant improvements in stability and enhanced running time of instruments~\cite{graham_high-throughput_2019, yin_petascale_2020, phelps_reconstruction_2021, xu_enhanced_2017} resulting in impressive advances in connectomics~\cite{motta_dense_2019, scheffer_connectome_2020, phelps_reconstruction_2021} as well as in cell biology~\cite{muller2021, parlakgul_regulation_2022, sheu_serotonergic_2022, weigel_er--golgi_2021, uwizeye_morphological_2021, musser_profiling_2021} and enabling even the reconstruction of whole organisms~\cite{vergara_whole-body_2021}.
Moreover, many of these datasets have been made publicly available in open-access repositories/platforms such as EMPIAR~\cite{iudin_empiar_2016}, CEM500K~\cite{conrad_cem500k_2021} and OpenOrganelle~\cite{xu_open-access_2021}.
The massive increase in data generation and accessibility has led to a surge in demand for analysis approaches.
In particular, this requires segmentation maps of the desired structures throughout the volume followed by analysis of i.e. volume fractions and/or investigation of spatial interactions. Until recently a major part of the segmentation work had to be done manually within software packages such as IMOD~\cite{kremer_computer_1996} which can become extremely time-consuming or even infeasible for larger datasets. Nevertheless, these approaches have lead to impressive results~\cite{noske_expedited_2008, wu_contacts_2017}.\\

The rise of machine learning methods has increased the possibility for automatic segmentation processes of a diverse set of organelle classes in large vEM images~\cite{kaynig2015,dorkenwald2017,buhmann2021,vergara_whole-body_2021,heinrich2021,spiers_deep_2021,gallusser_deep_2022}.
For example, Kaynig \textit{et al.}~\cite{kaynig2015} used random forest classifiers~\cite{breiman_random_2001} to semi-automatically segment neurons from ssEM images.
More recent, deep learning-based approaches based on U-Net~\cite{falk_u-net_2019} have shown impressive results on large scale organelle segmentation from FIB-SEM volumes, as part of the “Cell Organelle Segmentation in Electron Microscopy” (COSEM) project~\cite{heinrich2021} allowing pixel-wise segmentation for up to 35 cellular organelle classes.
However, deep learning-based methods typically rely on labor-intensive manual ground truth annotation efforts, and often require computational resources that smaller laboratories cannot afford. Furthermore, the bio-imaging community currently lacks protocols dealing with spatial analysis of 3D segmentation maps, such as the distribution of organelle-organelle distances, which prevents full exploitation of the raw data to obtain new biological insights. Finally, an effective visualization of the obtained volumetric segmentation maps is crucial for a holistic understanding of organelle distribution. At the same time, the fragmented landscape of visualization tools, that are often hard to learn, can be confusing for inexperienced users. 
In this protocol, we aim to address these issues by providing practical guidelines to perform the following key tasks on vEM images:  

\begin{itemize}
  \item Preparation of raw data  (\textbf{Procedure 1})
  \item Organelle-specific segmentation with a various set of methods (\textbf{Procedure 2})
  \item Spatial analysis of the segmentation maps (\textbf{Procedure 3})
  \item 3D visualization of segmentation maps (\textbf{Procedure 4})
\end{itemize}
In particular, we present detailed steps for organelle-specific segmentation of vEM datasets with user-friendly tools and give guidelines for which methods are most appropriate for each organelle class. We furthermore provide detailed steps to analyze the spatial interaction between the resulting 3D segmentation maps, thus enabling a comprehensive view of intracellular organelle interactions which is essential for understanding the complex cellular ultrastructure.
Finally, we present detailed instructions for generating 3D visualizations of the segmented organelles within their native cellular context and for creating accurate renderings for publication purposes and scientific outreach.

\figOverview

\subsection*{Development of the protocol}

We initially developed this protocol (\cf~\cref{fig:overview}) to investigate the interaction of microtubules and organelles in quasi-isotropic FIB-SEM volumes of primary beta cells with a voxel size of 4 nm~\cite{muller2021}.
Our aim was to analyze several complete beta cells in samples exposed to different metabolic stimuli (glucose concentrations). In particular, we focused on the organization of microtubules, their interaction with insulin secretory granules (SGs) and other organelles, as well as possible changes upon stimulation with glucose.
This required to perform 3D segmentation of various organelles such as microtubules, insulin SGs, nuclei, centrioles, mitochondria, Golgi apparatus, and the plasma membrane of each cell. These organelles strongly differ in many of their features: abundance in the cell, shape, and electron contrast.
Since some of them, such as the insulin SGs amount to several thousand vesicles per cell it became clear to us that full manual annotation is too time-consuming and cumbersome.
We also found that no single segmentation approach would perfectly match the specific features of all organelles simultaneously, and therefore chose dedicated methods that best suited a given organelle.
Specifically, we required: \emph{i)} Mostly automated segmentation approaches that \emph{ii)} do not demand a large computational infrastructure and labor-intense
 prior annotation, can be \emph{iii)} implemented in user-friendly software and iv) export file formats for segmentation results which can be easily shared between collaborators.\\

For full manual tracing of microtubules we chose \knossos~\cite{helmstaedter_high-accuracy_2011} since it allows memory-efficient annotation at the native pixel resolution, which is crucial for microtubules.
We used Microscopy Image Browser (MIB)~\cite{belevich_microscopy_2016} for manual segmentation of the plasma membrane and local thresholding of centrioles and nuclei. For more abundant and structurally conspicuous organelles such as mitochondria and (in a first attempt) insulin SGs we used autocontext classification~\cite{tu_auto-context_2010} implemented in ilastik~\cite{berg_ilastik_2019, kreshuk_machine_2019}. For more difficult tasks such as the Golgi apparatus we turned to deep learning using a U-Net architecture~\cite{falk_u-net_2019} implemented in CSBDeep~\cite{weigert_content-aware_2018}, whereas for refinement of insulin SG segmentation masks we used \stardist~\cite{weigert_star-convex_2020}. Finally, manual curation of the segmentation results was performed in ilastik or the Labkit plugin~\cite{arzt_labkit_2022} in \Fiji~\cite{schindelin_fiji_2012}.\\

We also found that a major hurdle for the proper analysis of 3D segmentation masks was the lack of protocols describing the analysis of complex spatial interactions. Analysis of organelle volumes is \eg possible in ilastik, IMOD or with the 3D object counter in \Fiji~\cite{bolte_guided_2006}. However, high resolution vEM segmentation masks allow the inspection of spatial interactions between organelles. This becomes especially interesting when investigating cells or tissues under different metabolic conditions as in our case~\cite{muller2021} and in other instances~\cite{parlakgul_regulation_2022} or when comparing states in health and disease~\cite{shrestha_integration_2022}. We therefore have devised a series of steps to perform these analyses which helped us to unravel the connectivity of microtubules to centrioles and the Golgi apparatus as well as their close interaction with insulin SGs.\\

The final step was the 3D rendering of the segmentation results for publication purposes and scientific outreach. We discriminate rendering for publication figures from 3D animations for outreach. For the former we used ORS (Object Research Systems) Dragonfly~(\url{https://www.theobjects.com/dragonfly/index.html}), which allows the loading of binary segmentation masks and the rendering of single images and animations of the cells. We used a more conservative rendering mode and a consistent color-code for these figures and videos. For 2D overlays of raw and segmentation data we used the \Fiji plugin 3Dscript~\cite{schmid_3dscript_2019}, which can also be used to generate 3D animations. For more engaging animations with shadows and special surfaces we used Blender~(\url{https://www.blender.org/}) which makes it necessary to convert the binary segmentation masks into meshes. This enables the generation of complex videos for public outreach. 

\subsection*{Advantages, applications, and comparison with other methods}

This protocol is aimed at researchers in the biology and medicine with only limited computational expertise who want to analyze vEM datasets of cells, tissues, or small organisms. 
Instead of relying on a single software, we provide a modular open-access workflow that does not require coding skills and enables users to pick out single methods for segmentation, analysis and rendering depending on their project.
In contrast to  multi-organelle segmentation approaches that use deep learning for all organelle classes and thus rely on the creation of massive amounts of user annotated ground truth for all organelles, our pipeline judiciously chooses organelle-specific segmentation methods to reduce unnecessary annotation effort. This strategy enables to achieve reasonable results early in the project for certain organelles, so that more time can be spend on difficult segmentation tasks.
Integration of spatial analysis and 3D rendering into easily executable solutions with Album~\cite{albrecht_album_2021} allows users to best address complex organelle interactions within cells and to obtain novel biological insights from their vEM datasets. Furthermore, the simple but effective Blender workflow enables the complete inner life of cells to be visualized and the reconsideration of our perceptions derived from 2D images. \\

The use case that we primarily address in this protocol is organelle segmentation in FIB-SEM volumes of beta cells. However, we designed the segmentation, analysis and rendering workflows in a way that allows the simple adaptation to other cell types and vEM modalities. Possible research questions include investigating the interaction of organelles with microtubules, contact sites between organelles such as mitochondria and the endoplasmic reticulum (ER) and changes of organelle morphology and interactions under different metabolic stimuli or disease conditions. We also aim to help scientists to decide which segmentation workflow suits their target structure best and to find a time- and labor-efficient strategy. Furthermore, the protocol shall serve as guide to vEM analysis for scientists that do not have extensive experience with EM yet. We originally developed the protocol for analyzing isotropic FIB-SEM volumes containing complete cells. However, the majority of the presented methods can also be applied to anisotropic datasets derived from ssTEM or SBF-SEM imaging or smaller volumes comprising only a few serial sections or single tomograms.
Because of its modular design, our approach is complementary to other vEM segmentation and analysis tools (\eg Amira, Empanada~\cite{conrad_instance_2023}, workflows provided by ZeroCostDL4Mic~\cite{von_chamier_democratising_2021}, pre-trained models of the BioImage Model Zoo~\cite{ouyang_bioimage_2022}) as these can be integrated in our protocol at the appropriate steps.
Therefore, our strategy can serve as a time- and labor-efficient framework for vEM analysis.

\subsection*{Limitations}

Currently, we do not present strategies for all organelle classes such as ER, endosomes, or lysosomes. Recent publications describe time-efficient tools to segment these structures~\cite{gallusser_deep_2022} as well as the nuclear envelope~\cite{spiers_deep_2021}, and mitochondria on an instance level~\cite{conrad_instance_2023}.
However, these tools could be used at the appropriate steps within our procedures.
Additionally, the completely manual annotation of microtubules presented in our protocol is currently very time consuming and posed a bottleneck for our project.
There are approaches using template-matching~\cite{weber_automated_2012} or deep learning~\cite{eckstein_microtubule_2020} that might help automatizing this task in the future.
Our segmentation procedure of complex organelles via U-Net is currently not implemented as a GUI based software solution. The recently released tool DeepMIB~\cite{belevich_deepmib_2021} as part of MIB provides such an interface and thus might be implemented into our pipeline as demonstrated for local threshold segmentation.\
Furthermore, vesicle segmentation with \stardist should work similarly well for other large dense core vesicle types whose ultrastructure resembles SGs (\eg chromaffin granules).
For other, more distinct, vesicle types (\eg synaptic vesicles) alternative approaches, such as in~\cite{kaltdorf_automated_2018} might be more appropriate.\\

Our protocol was tested on quasi-isotropic FIB-SEM data and therefore some of the segmentation tools described might work less well for anisotropic data modalities such as ssTEM or SBF-SEM. There it might be necessary to switch to tools developed specifically for these type of data, such as CDeep3M~\cite{haberl_cdeep3mplug-and-play_2018}, which can be included into our workflows.

Finally, our current data format (TIFF) for exporting most of the segmentation results is sub-optimal since it does not allow for a very efficient saving and reading of the data.
Hence switching to hierarchical data-formats (\eg HDF5~\cite{koranne_hierarchical_2011}, N5~\cite{saalfeld_saalfeldlabn5_2022} or ZARR~\cite{miles_zarr-developerszarr-python_2020}) will be beneficial in the future (with the caveat that not all segmentation tools currently support these formats).
Therefore, we only used HDF5 for segmentation in ilastik and kept TIFF for the other segmentation tasks. As a result, our workflow might struggle for excessively large vEM datasets as used \eg in connectomics. However, we used N5 for the spatial analysis and recommend to convert already the raw data to one of the aforementioned file formats whenever possible. Designing workflows with interconnected components fully working on hierarchical data formats would be beneficial for improving the scalability of the protocol.\\

\figSegmentation

\subsection*{Experimental Design}

In the following, we give some methodological background to the three main subtasks that we will later describe in the protocol, \ie organelle segmentation, spatial analysis, and visualization. 

\subsubsection*{Organelle Segmentation}

Before beginning to annotate structures it is important to have an overview on different methods and to decide which strategy might fit best the organelle to be segmented. In~(\cref{fig:Segmentation}) we list segmentation strategies according to organelle features that shall help with making this decision.

\begin{protobox}[t!]{How to choose a segmentation method}
Different organelle classes pose different challenges to an automatic segmentation method: A method that accurately segments mitochondria might not be a suitable choice for delineating the Golgi apparatus and vice versa. 
Knowing which segmentation method is best suited for each organelle class is thus critical  for both the accuracy of the final results and to avoid time-consuming annotations in cases where they would not be required.  
In general, this choice depends on structural features of the organelle such as typical image contrast, organelle abundance, and morphological complexity.
Rare and contrast-rich organelles such as centrioles or nucleus, for example, often can be segmented with relatively simple tools that do not require manual annotation effort, such as local thresholding.
On the other hand, abundant but well separated organelles, such as mitochondria typically are best segmented  with machine-learning based methods such as random forests or convolutional neural networks (CNNs) which in turn require a certain amount of manual annotations.
For this class of organelles, we recommend to start with random forest classifiers (such as provided \eg by ilastik) that allow for quickly judging wether more advanced methods such as a U-Net based semantic segmentation with CNNs are required.
Finally, some organelle classes may present unique challenges, such as the extreme dense and crowded distribution of SGs that are hard to properly separate.
For such situations problem-specific applications might exists, such as the shape-aware method \textsc{StarDist} for the detection of crowded spherical objects. 
In~\cref{fig:Segmentation} we summarize these guidelines for when to use manual, semi-automated or machine-learning-based segmentation tools, which shall serve as an inspiration for any new segmentation endeavor.
\end{protobox}

\paragraph{Semi-Automatic Local Thresholding and Manual Segmentation:}

For some organelles classes with low abundance (\eg centrioles) segmentation via locally adjusted thresholding at each instance can already be sufficient and at the same time relatively quick to perform.
Local thresholding works by manually segmenting the area very close to the target organelle followed by black-and-white thresholding within this defined area.
We use this strategy for nuclei and centrioles and chose MIB as a tool, as it allows to export segmentation results in various formats, such as Amira AM file or TIFF.
For other classes (\eg plasma membrane) such simple approaches might fail to detect the structure of interest, and laborious manual segmentation might become necessary.
Manual segmentation is basically done by tracing the outlines of the target organelle with a pen or brush tool in every z-section of the stack. It can be facilitated by interpolation in the z-dimension every five or ten slices. 
We again used MIB for manual segmentation of the plasma membranes of the cells.

\paragraph{Random Forest Segmentation:}  
Random forest based segmentation methods~\cite{breiman_random_2001, schroff2008} use an ensemble of decision trees to classify each pixel based on a set of predefined image features.
Random forest classifiers need only sparse (\eg scribble) annotations and are fast to train, making them the classical machine-learning method of choice for initial segmentation experiments without the need for excessive manual annotation.
Several software packages, such as MIB, \Fiji (with the plugins trainable weka~\cite{arganda-carreras_trainable_2017} or Labkit~\cite{arzt_labkit_2022}), and ilastik provide random forest classifiers.
Ilastik additionally implements a cascaded random forest classifier (\emph{autocontext}) which can achieve better results than the simple pixel classification, but includes two rounds of annotation and training. In the first round the user needs to annotate many different organelles (\eg nucleus, vesicles, mitochondria, ER, Golgi, ribosomes). After training only background and one organelle (\eg mitochondria) are labeled. The second round of training will enable better results compared to a single-step pixel classification. We used this approach for segmentation of mitochondria and to generate ground truth data for insulin SGs and the Golgi apparatus.

\begin{protobox}[t]{How to generate ground truth}
  Machine-learning based segmentation methods (such as random forest and deep learning) require training data that consists of raw images and corresponding  annotations (\emph{ground truth}).
  For most random forest methods sparse (scribble) annotations are sufficient for good segmentation result, whereas many deep-learning based segmentation methods require fully labeled ground truth images.
  As typically only a small part of the full raw images can be annotated with reasonable effort (\eg 5-10 subregions of \sizethree{128}{128}{128}~pixels), the selection of which subregions to annotate is crucial for the performance of the trained model.
  In particular, these subregions should cover as best as possible the image variability seen in the full dataset.
  That means that you should not exclusively provide labelings of your organelle of interest, but also regions without it, such as extra-cellular space or regions rich in other organelles/structures. We also advise to label several small crops of the raw data and not a large chunk of the dataset. One should cut several small structurally diverse volumes ideally out of several raw datasets to ensure a good final segmentation result. To generate proper ground truth a high level of expert knowledge is necessary. Classic EM textbooks and recent publications such as~\cite{park_amira_2022} can help to decide how to distinguish organelles.
  As the generation of pixel-accurate annotation masks from scratch can be quite time-intensive, we generally recommend to create preliminary ground truth with a fast and easy to use method first (\eg ilastik) and then refine and curate them manually in a second step.
  This way, most of the annotation effort is spend on hard structures and facilitates the process of ground truth generation and allows non-expert users to faster get an idea on structural features of the different organelles.
\end{protobox}

\paragraph{Deep Learning:}
Deep learning based on multi-layered convolutional neural networks (CNNs) has become the most prevalent method for challenging image segmentation problems in the life sciences~\cite{hallou_deep_2021}.
Commonly used CNN architectures are able to capture high and low resolution image semantics while containing millions of parameters that are learned from training data without the need for handcrafted features.
Although this can result in very expressive and powerful segmentation models, the amount of training data required for good segmentation results is typically larger than for less expressive methods (\eg random forests) and thus the manual annotation effort is often the bottleneck in those workflows.
The most widely used architecture for microscopy image segmentation is U-Net~\cite{falk_u-net_2019}. Moreover, network ensembles also have led to good results~\cite{haberl_cdeep3mplug-and-play_2018, shaga_devan_weighted_2022}.

\paragraph{Shape-aware deep learning:}

For some organelle classes it can be beneficial to exploit the structural priors they might possess, \eg for SGs or other vesicles that are typically spherical in shape. In these cases \emph{shape-aware} deep learning methods that directly incorporate these priors in their prediction targets can be effective. Examples of these include \stardist~\cite{weigert_star-convex_2020}, SplineDist~\cite{mandal2021}, Cellpose~\cite{stringer2021}, or \emph{local shape descriptors} (LSD)~\cite{sheridan2022}.

\paragraph{Skeleton Tracing of Filamentous Structures:} Components of the cytoskeleton such as microtubules are filamentous structures that can be represented as skeletons with a defined thickness. Since microtubules have an outer diameter of 25 nm it was in our case necessary to use the full resolution of the dataset for annotation. \knossos uses a special file-format enabling the loading of only a small region surrounding the currently viewed area which makes it very memory-efficient. Skeleton tracing is then performed by going through the volume and setting points throughout single microtubules from start to end which are then connected by a line. These skeleton traces are then analyzed within the album workflow.

\subsubsection*{Volume and Spatial Analysis}
Analysis of the segmented data usually starts with calculating volumes and volume fractions of the different organelles with respect to the whole cell or raw volume. These calculations can be performed by the several image analysis programs such as IMOD, \Fiji, MIB or Amira. These data help to address the heterogeneity between cells and to investigate changes in organelle volumes, i.e., upon metabolic changes or in disease conditions. Precise segmentation masks also allow the analysis of distances between organelle classes either directly out of the binary masks or after creating 3D meshes. These approaches were first established for analyzing microtubule interactions in electron tomograms of mitotic spindles~\cite{mcdonald_kinetochore_1992} and organelle interactions within the Golgi region of a beta cell line~\cite{marsh_organellar_2001}. The latter study also introduced the comparison between observed and random distributions in order to make assumptions on regulated processes of organelle interaction - a method we have adopted for our original study. These kind of calculations enable the investigation of complex interactions between organelles and the definition of subsets of an organelle based on its connectivity. Tools are partially implemented in IMOD, Amira and in the Neuromorph toolset for blender~\cite{jorstad_neuromorph_2015, jorstad_neuromorph_2018}, however they are not yet widely adopted. We generated distance transformations and to calculate distances and connectivity between them with ImageJ2~\cite{rueden_imagej2_2017} and ImgLib2~\cite{pietzsch_imglib2generic_2012}. Processing and plotting of these data was performed with Python-based tools~\cite{harris_array_2020, hunter2007, virtanen_scipy_2020, mckinney2010}. In this protocol these workflows are executed via Album.

\subsubsection*{Visualization}

A successful 3D segmentation allows for 3D rendering of the generated organelle masks to visualize all or a selection of organelles in their proper context. There is a number of tools available that allows the direct rendering of the original masks such as the 3D viewer~\cite{schmid_high-level_2010} or 3Dscript~\cite{schmid_3dscript_2019} in \Fiji or in Dragonfly. Dedicated 3D software tools such as blender usually require conversion of the binary masks into 3D meshes. Before starting the 3D visualization you should define a color scheme for your organelles with the help of online tools such as \url{https://coolors.co/}. Ideally, this should be colorblind-friendly or at least structurally similar organelles such as different vesicle types should have distinct colors. You can check this with the Simulate Color Blindness plugin in \Fiji.  

\begin{protobox}[t]{How to effectively visualize results}
As an integral part of basic research in medicine, biomedical image datasets can establish links to personal experiences of individuals with health or diseases. The 3D modeling and rendering software Blender is exceptionally well suited to tell the story behind a 3D dataset. Blender is extremely versatile within and beyond object material adjustment, illumination, and animation. It is open source, freely available and fully scriptable, enabling us to simplify complex routines like importing specific formats, adjusting the camera to frame all imported objects and assigning materials to the objects automatically. Scientific images are usually represented as pixel data - as a grid of points where each point has a specific value. In order to import such data into Blender, it first needs to be converted into a geometric structure by distinguishing between background and foreground pixels and generating a mesh of vertices along the border of the foreground, representing an object in 3D. Our protocol includes a routine for this purpose.
\end{protobox}

\newpage

\section*{Material}

\subsection*{Equipment}
\subsubsection*{Hardware requirements}

For training deep learning models (Procedures 2c and d) a \textsc{Nvidia} GPU is indispensable. Also sufficiently high memory (at least 32 GB) is necessary to handle raw data and segmentation masks.
For the original study we used:
\begin{itemize}
\item   ThinkPad P52 with 32 GB RAM and a \textsc{Nvidia} Quadro P3200 GPU with 6 GB VRAM with Windows10
  \item  Workstation with 128 GB RAM and a \textsc{Nvidia} RTX 2080 with 8 GB RAM with Ubuntu 18.04
\end{itemize}

The analysis and visualization procedures were additionally tested on:
\begin{itemize}
    \item Desktop Computer with 32 GB RAM and a NVIDIA GeForce RTX 3060 with 12 GB VRAM with Windows10
    \item ThinkPad P14s with 46 GB RAM and a NVIDIA T500 with 4 GB VRAM with Ubuntu 20.04
\end{itemize}

\subsubsection*{Datasets}
\begin{itemize}
    \item High resolution raw data: \url{https://openorganelle.janelia.org/}
    \item Example data for deep learning and spatial analysis: \url{https://desycloud.desy.de/index.php/s/Cib6YP642qaPxqs?path=\%2F}
\end{itemize}

\subsection*{Software}
All software used in this protocol can be found in~\cref{table:2}.
\begin{table*}[h]
  \centering
  \footnotesize
  \def\arraystretch{.8}
\begin{tabular}{ p{3cm} p{8cm} p{4cm}} 
 
  \toprule
  \textbf{Tool} & \textbf{Purpose} & \textbf{Link}\\
\midrule
\textbf{Album} & Decentralized distribution platform for solutions to specific scientific problems. Here, we run solutions from the Helmholtz Imaging Solutions catalog for spatial analysis of vEM segmentation data and for 3D rendering. & \url{https://album.solutions}, \url{https://gitlab.com/album-app/catalogs/helmholtz-imaging} \\ 
\midrule

\textbf{Blender}  & Software for generating 3D renderings. Here, we use it to render our segmentation masks. & \url{https://www.blender.org/} \\
\midrule

\textbf{\cellsketch} & Tool suite for displaying and analyzing cell components, utilizing BigDataViewer, ImageJ, VTK, and Blender. &  \url{https://github.com/betaseg/cellsketch}\\
\midrule
 
\textbf{Conda} \\ \eg Miniconda & Python distribution for running album and other python-based tools of the workflow. & \url{https://docs.conda.io/en/latest/miniconda.html} \\

  \midrule
  
\textbf{Python Packages} & Additional packages that can be installed with \texttt{pip} & \\[.2cm]

  \quad \texttt{csbdeep} & Library providing generic deep learning models, \eg U-Net  & \url{https://github.com/CSBDeep/CSBDeep/}\\ 

  \quad \texttt{stardist} & Library for 2D/3D shape-aware segmentation of spherical objects such as nuclei or vesicles & \url{https://github.com/stardist/stardist}\\ 

\midrule
 
\textbf{\Fiji} & Software for bio-image visualization and analysis. Here, we use several plugins for our workflow. & \url{https://imagej.net/software/fiji/}\\ 
\midrule

\textbf{\Fiji plugins} &   & \\[.2cm]

\quad 3Dscript & Plugin for language-based creation of animations of image data. Here, we use it to generate overlay images of raw and segmentation data. & \url{https://bene51.github.io/3Dscript/}\\

\quad BigDataViewer & Plugin for loading large image files. & \url{https://imagej.net/plugins/bdv/}\\

\quad Ilastik& Plugin for converting TIFF to HDF5 files in FIJI and vice versa & \url{https://github.com/ilastik/ilastik4ij}\\

\quad Labkit & Plugin for annotation and random forest classification. Here, we use it for annotation and cleaning of ground truth data. & \url{https://imagej.net/plugins/labkit/}\\
\midrule

\textbf{Ilastik} & Software for image segmentation. Here, we use it for segmentation with random forests. & \url{https://www.ilastik.org/}\\
\midrule

\textbf{IMOD} & Image processing software for reconstruction and segmentation of electron tomography datasets. Here, we use its programs to cut and bin the original vEM data. & \url{https://bio3d.colorado.edu/imod/}\\
\midrule

\textbf{\knossos and \knossos cuber} & Software for opening and annotating large vEM datasets. Here, we use it for skeleton tracing of microtubules. \knossos cuber converts volume EM files to cubed files readable by \knossos. & \url{https://knossos.app/}, \url{https://github.com/knossos-project/knossos_cuber}\\
\midrule

\textbf{Microscopy Image Browser} & Software for segmentation of electron microscopy data providing a large variety of tools. Here, we use it for manual and local thresholding segmentation of volume EM datasets. & \url{http://mib.helsinki.fi/}\\
\midrule

\textbf{ORS Dragonfly} & Software for segmentation, visualization and analysis of 3D image data. here, we use it for generating simple renderings for publication figures. & \url{https://theobjects.com/dragonfly/index.html}\\
\midrule

\textbf{Visualization Toolkit (VTK)} & Software for manipulation and visualization of scientific results. Here, we use it to evaluate meshing for 3D rendering. & \url{https://vtk.org/}\\

 \end{tabular}
\caption{Software used in this protocol}
\label{table:2}
\end{table*}

 \newpage

\section*{Procedures}

\subsection*{Procedure 1 - Preparation of Raw Data}
\figRaw

This protocol starts with an aligned vEM stack. Instructions on alignment can be found elsewhere~\cite{cardona_trakem2_2012, hennies_amst_2020}. You should take care that the voxel dimensions of the vEM stack have been properly calculated~\cite{hanslovsky_image-based_2017}. Also, it might be helpful to perform a denoising step on your dataset. There are various tools available \eg DenoisEM~\cite{roels_interactive_2020} or Noise2Void~\cite{krull_noise2void_2019}. Before starting to segment, you should determine if you need to work on the full volume or if it is possible to cut certain regions of interest (ROI), i.e. each ROI containing a cell~(\cref{fig:raw}). Furthermore, you might not need the full resolution of the original dataset to segment organelles such as mitochondria or nuclei. Binning by a factor of 2 or 4 might make sense as suggested in~\cite{perez_workflow_2014, hoffman_correlative_2020} and our own work~\cite{muller2021}. This will save a substantial amount of hard drive and memory space. We only used the full resolution for segmentation of microtubules. You should cut ROIs before binning. However, if your original stack is too big for opening it on your computer you can bin it before you cut ROIs. You have to be extremely careful to note the correct coordinates of the ROIs in the binned volume and transfer it to the full resolution raw stack. You can then cut high resolution ROIs with the \texttt{trimvol} program as part of IMOD.

\subsubsection*{Procedure 1a - Cutting ROIs containing complete cells - TIMING: 10-30 min}
\begin{steps}
\item Open the full resolution image stack in \Fiji.\\
\textbf{? TROUBLESHOOTING}
\item Define the ROI in xy with the "Rectangle" tool followed by Image \menu Crop.\\ Afterwards go through the resulting stack in z and define start and end of the cell. Cut the cell in z by Image \menu Stacks \menu Tools \menu Slice Keeper and define "First Slice" and "Last Slice". Make sure to use "Increment" 1 and press OK. Save the resulting stack as TIFF file.
\end{steps}

\subsubsection*{Procedure 1b - Binning of subvolumes - TIMING: 5 min }
\begin{steps}
\item Open the subvolume in \Fiji
\item Go to Image \menu Transform \menu Bin and define X, Y and Z "Shrink Factor". You also need to choose the "Binning Method" (we recommend Average). Press OK and save the result as TIFF file.\\
  Alternatively you can use \texttt{binvol} as part of IMOD for binning. To do this open the terminal and type binvol followed by the binning factor, your original file and the destination and file name of the result.
\end{steps}

\subsubsection*{Procedure 1c - Preparation for \knossos - TIMING: 10-30 min}
For skeleton tracing in \knossos you will need to convert your full resolution vEM dataset with \knossos cuber. 
\begin{steps}
\item Install the program by following the installation instructions on \url{https://github.com/knossos-project/knossos_cuber/blob/master/README.md}.
\item Use the graphical user interface to define the "source directory", the "target directory" and "source format". 
\item After running the program you can open the result in \knossos.
\end{steps}

\subsubsection*{Procedure 1d - Preparation for ilastik - TIMING: 10 min}
Ilastik can open various file formats. However, for optimal performance you should convert your original file to HDF5. You can do this with the ilastik plugin in \Fiji. 
\begin{steps}
\item Open your vEM stack in \Fiji.
\item Go to Plugins \menu Ilastik \menu Export to HDF5. 
\item Define the "Export Path" and press OK.
\end{steps}

\subsection*{Procedure 2 - Segmentation}
The following paragraph will describe workflows for organelle-specific segmentation organized by increasing difficulty including methods ranging from manual segmentation to deep learning.

\subsubsection*{Procedure 2a - Local Thresholding and Manual Segmentation - TIMING: 45-120 min}
\figMIB

\begin{steps}
\item Open Microscopy Image Browser.
\item Open your raw dataset by clicking the "..." button and navigate to the folder where the file is saved in. Press OK and click on the file in the left sidebar~(\cref{fig:MIB}a).
\item Navigate to your organelle of interest (\eg centrioles).
\item With the "Brush" tool draw a line close to (for local threshold) or precisely on the organelle. You can fill the annotation by pressing the F key.
\item Annotate every 5th or 10th slice depending on the voxel size of the dataset. Press the I key for interpolating between annotated slices.
\item After finishing manual annotation go to Selection \menu ... \menu Mask \menu 3D \menu Current stack(3D) \menu Add. Now the outline of your annotation will appear in magenta.
\item Switch from the Brush tool to "BW Thresholding". Make sure to check the box "Masked area" to only perform thresholding within the mask and also check the "3D" box.
\item \textbullet \textbf{CRITICAL STEP} Change the high and low thresholds until you see a reasonable labeling of your target organelle in green.
\item Convert your labeling into a mask as described in 6) but replace the Current stack.
\item Export the mask as AM or TIFF file by clicking Mask \menu Save mask as.
\item Import the file into \Fiji.
\item It might happen that you do not see the annotated area after import into \Fiji. In this case go to Process \menu Binary \menu Convert to Mask. Select "Huang" for binarization and click OK.
\item[] \textbf{? TROUBLESHOOTING}
\item \textbullet \textbf{CRITICAL STEP} For plasma membrane segmentation and for the nucleus (especially if you want to perform spatial analysis later) you should do a precise annotation without local thresholding. In these cases export the mask as in step 10) after labeling.
\end{steps}  

\subsubsection*{Procedure 2b - Random Forests - TIMING: 4-24 hrs} 
\figIlastik

\begin{steps}
\item Open ilastik and create a new project by clicking on "Autocontext (2-stage)". Define a file name for the project and save it.
\item Load the raw data in "1. Input Data" by clicking +Add New.. \menu Add separate Image(s).. and chose the HDF5 file you want to open. Ilastik will then display the file in xy, yz and xz view.
\item As next step select "2. Feature Selection" to select pixel features. Make sure to have 3D selected especially for isotropic data. For datasets with a huge difference between x, y and z-values it can make sense to select 2D features. More features will increase the likelihood for better results, but also require more computational resources. Press OK after selecting the features.
\item In the next step you will perform a first round of training~(\cref{fig:ilastik}a). Click in "3. Training" and define labels for your organelles. In this first round you should label more than two different organelles, ideally as many as you can discriminate. You can add new labels by clicking "+ Add Label", change colors of the labels by right click on the color and change label names by left double click. After defining labels (in our case: vesicles, mitochondria, chromatin and background) perform sparse labeling of the structures with the "Brush Cursor" tool for all labels. Press "Live Update" to see the probability map for each label~(\cref{fig:ilastik}b). In this first step this does not have to be perfect. However, you can improve the results by providing more annotations.
\item By clicking on "Suggest Features" you can run an automated feature selection that can help to optimize the manual selection you did in the previous step. Make sure to click Project \menu Save Project in between the steps.
\item If you are satisfied with the first round of training you may continue to "4. Feature Selection" to select features for the second training. After feature selection and clicking OK proceed to "5. Training". Define labels similar to the first training step. However, now you only provide annotation for your desired organelle (in our case mitochondria) and background~(\cref{fig:ilastik}c). Press "Live Update" and provide more annotations until you are satisfied with the result~(\cref{fig:ilastik}d). In this step you can also change features or get suggestions by clicking on "Suggest Features".
\item If the probability map for your target organelle looks reasonable you can proceed to "6. Prediction Export". In the "Export Settings" you can decide which file you wish to export. We will export "Probabilities Stage 2" since we later want to perform object detection. Click on "Choose Export Image Settings...", set the file format (HDF5) and give a file name. Click OK and then click "Export".
\item[] \textbf{? TROUBLESHOOTING}
\item In order to generate binary masks needed for analysis you can either export the "Simple Segmentation Stage 2" files after pixel classification or start a new project "Object Classification [Inputs: Raw Data, Pixel Prediction Map]".
\item Load raw data and the corresponding probability map.
\item Change thresholds until you are satisfied with the binary output.
\item In "3. Object Feature Selection" you can select the features you want to be computed.
\item In "4. Object Classification" you need to once label the organelle and press "Live Update". You will then see all mitochondria labeled~(\cref{fig:ilastik}e) and can proceed to "5. Object Information Export".
\item Export the "Object Predictions" as HDF5.
\item You can import the HDF5 files into \Fiji with the ilastik plugin by going to Plugins \menu Ilastik \menu Import HDF5 and select the file. It might be necessary to perform an additional binarization step as described for importing files from MIB in 2a, step 12. Afterwards you can save the file as TIFF.
\item[] \textbf{? TROUBLESHOOTING}

\end{steps}

\subsubsection*{Procedure 2c - Deep Learning for complex and/or convoluted organelles - TIMING: 12-36 hrs} 
For organelles that are highly convoluted and distributed throughout the whole cell, such as the Golgi apparatus or the ER, a supervised deep learning based method is often the most efficient choice. As semantic segmentation network we use a U-Net as a robust and popular architecture for bio-images. For supervised deep learning you need to provide precisely annotated ground truth (see box). We found that it can help to generate rough segmentation masks with Autocontext in ilastik (see procedure 2b) that can be cleaned in ilastik or Labkit similar to the data curation step in procedure 2f.

\begin{demodata}{Segmentation of the Golgi apparatus}

  If you want to try out Procedure 2c without gathering your own data first, we provide example demo data for segmentation of the Golgi apparatus.

  \begin{center}
  \includegraphics[width=.4\linewidth]{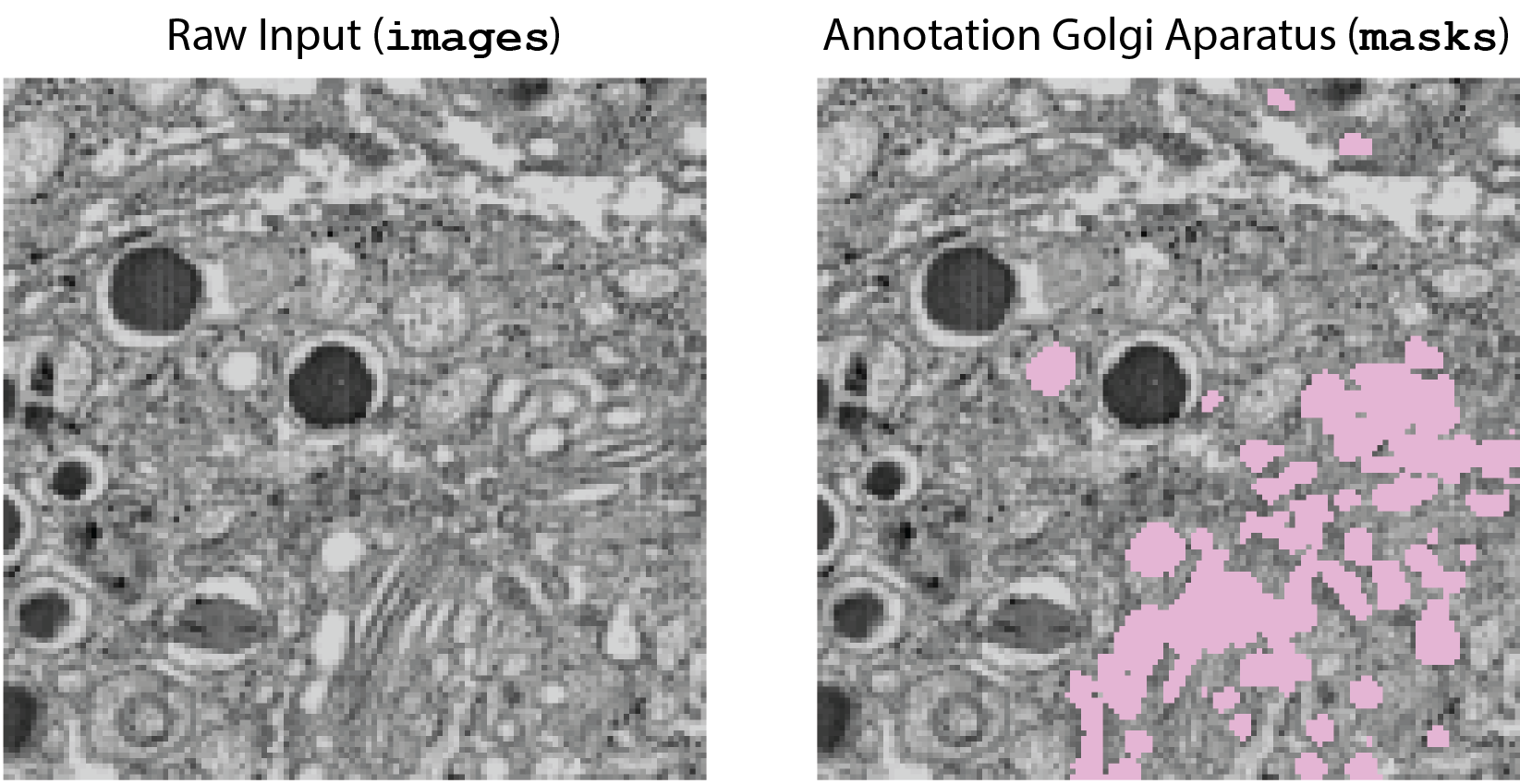}
  \end{center}

  You can download the data via
  \begin{itemize}
    \footnotesize
  \item[\$] \lstinline{wget https://syncandshare.desy.de/index.php/s/FikPy4k2FHS5L4F/download/data_golgi.zip}
    \item[\$] unzip data\_golgi.zip
  \end{itemize}

  The data consists of 9 FIB-SEM subvolumes of size \sizethree{128}{128}{128} pixels and corresponding dense label masks of annotated Golgi apparatus.
  At  \url{https://github.com/betaseg/protocol-notebooks} we additionally provide a corresponding \texttt{jupyter} notebook that demonstrates how to train a 3D U-Net on this data and how to use it to predict segmentation maps on new stacks. 

\end{demodata}

If you want to generate ground truth from scratch with Labkit proceed as follows:
\begin{steps}
  \item Open your raw dataset in \Fiji.
  \item Select square ROIs with the "Rectangle" tool. Click on Edit \menu Selection \menu Specify and type in the identical x and y dimensions. We recommend \sizetwo{128}{128} or \sizetwo{256}{256} pixels. Press OK. Press Image \menu Crop to crop the image in x and y.
\item To crop the resulting stack in z go to Image \menu Stacks \menu Tools \menu Slice Keeper and define the number of slices you want to keep (128 or 256). Make sure to set the Increment to 1. Press OK and save the resulting stack.
\item Repeat this to generate approximately 20 sub-volumes ideally of various different raw volumes.
\item Open one of the volumes in \Fiji and change z-dimension to time [t] by clicking Image \menu Properties and switch t- and z-values. Click OK and save the file.
\item Open the file in Labkit by clicking Plugins \menu Labkit \menu Open Current Image in Labkit.
\item Start annotation of your organelle of interest with the "Draw" tool. You can navigate through the dataset in z with the left and right arrow keys. After carefully labeling all structures either with one foreground label or with several distinct labels save your labeling as TIFF file.
\end{steps}

Alternatively, you can generate ground truth with ilastik by following these steps:
\begin{steps}[resume]
\item Prepare a first preliminary segmentation of the vesicles with Autocontext in ilastik (as shown in Procedure 2b) and export the resulting binary mask into \Fiji~(\cref{fig:DL}).
\item Out of this mask prepare crops with dimension of \sizethree{128}{128}{128} pixels and if necessary correct the labeling in Labkit (as described before).
\item Prepare 20 crops of corresponding raw images and labels masks and save them as TIFF files within two folders \texttt{data/train/images} and \texttt{data/train/masks}.
\end{steps}

Use the ground truth label files and corresponding raw files to train a semantic segmentation model as described in the notebook \url{https://github.com/betaseg/protocol-notebooks/blob/main/unet/run_unet.ipynb}. Specifically:
\begin{steps}[resume]
\item Load the corresponding raw images and label masks and define appropriate data augmentations. A good default strategy is to use intensity scaling, flip/rotations, and elastic deformations along the lateral dimensions.
\item Define a U-Net model with the default hyperparameters. Adjust the pooling factor to the anisotropy of the data (\eg \lstinline{unet_pool_size=(2,2,2)} for isotropic data,  or \lstinline{unet_pool_size=(2,4,4)} for data with twice as large axial  than lateral pixelsize).
\item \textbullet \textbf{CRITICAL STEP} Train the model and monitor the training and validation loss in \texttt{tensorboard}. The training loss should gradually decrease, while the validation loss should plateau after 100-150 epochs. If the validation loss starts increasing heavily for later epochs, then the model \emph{overfits} which needs to be avoided.
\item[] \textbf{? TROUBLESHOOTING}
  \item Use the trained model to predict on new (full) raw stacks using the prediction code within the notebooks at \url{https://github.com/betaseg/protocol-notebooks/blob/main/unet/run_unet.ipynb}. Depending on the size of the stack, the prediction step may take 0.2-1h.
\end{steps}

\figDL

\subsubsection*{Procedure 2d - Shape-aware Deep Learning for densely packed vesicles -   TIMING: 12-36 hrs}
For very abundant and densely packed star-convex (roundish) objects such as SGs or neuronal vesicles \stardist is a well suited shape-aware segmentation method. \stardist  directly predicts polyhedronal shape approximations for each object and typically yields well separated label maps even for very crowded scenarios. 

\begin{demodata}{Instance segmentation of insulin SGs}
  If you want to try out Procedure 2d without gathering your own data first, we provide example demo data for a segmentation task of densely packed SGs.

  \begin{center}
  \includegraphics[width=.4\linewidth]{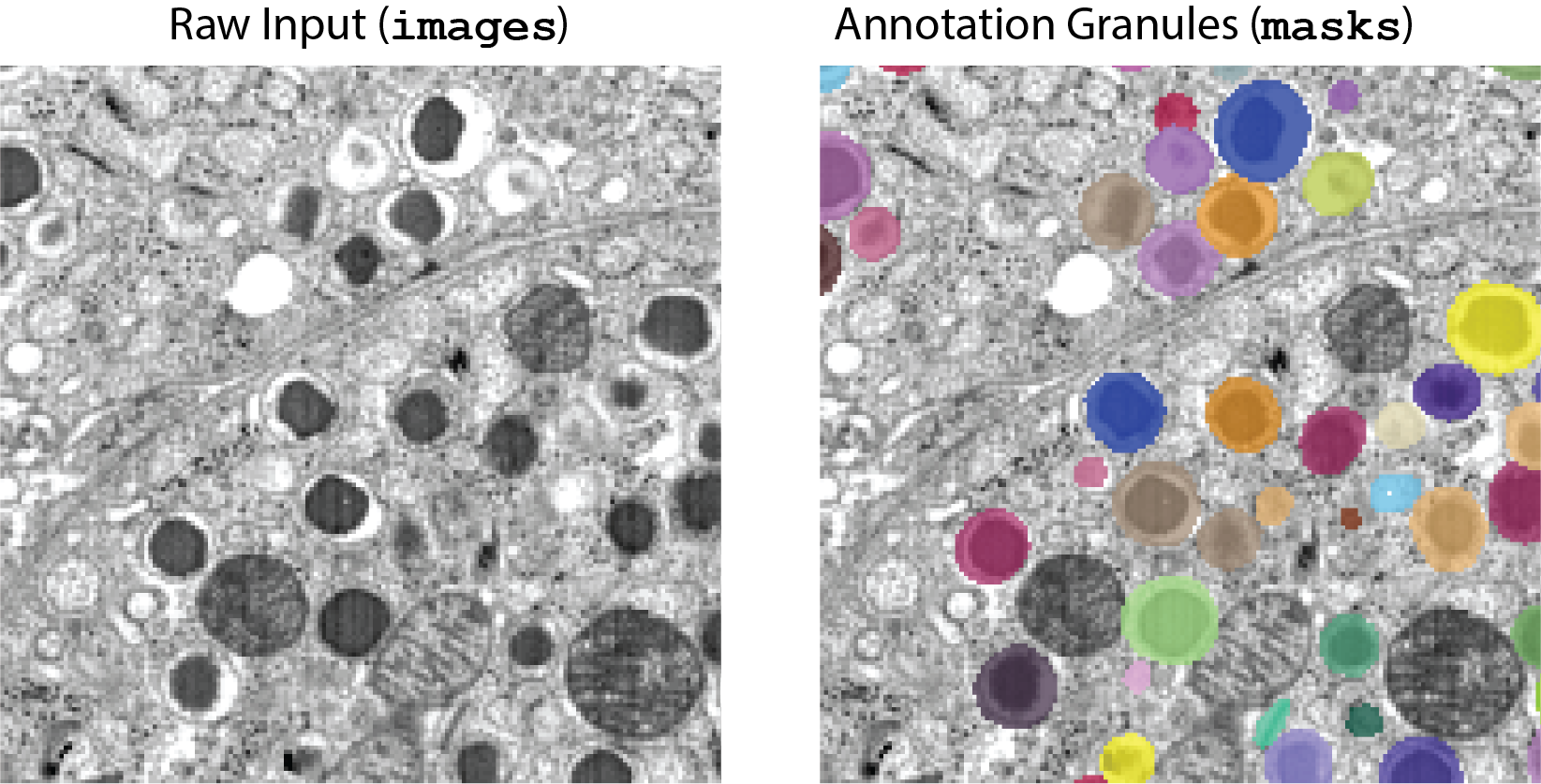}
  \end{center}
  
  You can download the data via
  \begin{itemize}
    \footnotesize
  \item[\$] \lstinline{wget https://syncandshare.desy.de/index.php/s/5SJFRtAckjBg5gx/download/data_granules.zip}
    \item[\$] unzip data\_granules.zip
  \end{itemize}

  The data consists of 5 FIB-SEM subvolumes of size \sizethree{128}{128}{128} pixels and corresponding dense label masks of annotated SGs.
  At  \url{https://github.com/betaseg/protocol-notebooks} we additionally provide a corresponding \texttt{jupyter} notebook that demonstrates how to train a 3D \stardist model on this data and how to use it to predict SG segmentation maps on new stacks. 
  
\end{demodata}

\begin{steps}
\item Prepare a first preliminary segmentation of the vesicles with Autocontext in ilastik (as shown in Procedure 2b) and export the resulting binary mask into \Fiji~(\cref{fig:DL}).
\item Out of this mask prepare crops with dimension of \sizethree{128}{128}{128} pixels and if necessary correct the labeling in Labkit (as described in Procedure 2c).
\item Prepare 20 crops of corresponding raw images and labels masks and save them as TIFF files within two folders \texttt{data/train/images} and \texttt{data/train/masks}.
\item Use these crops as ground truth to train a StarDist 3D model with the default settings using the provided notebooks at \url{https://github.com/betaseg/protocol-notebooks/blob/main/stardist/run_stardist.ipynb}.
  Note that training requires a workstation with a \textsc{Nvidia} GPU and will take up to 6-8h.
\item \textbullet \textbf{CRITICAL STEP} The model performance during training can be monitored by observing the training/validation loss in \texttt{tensorboard}. The training loss should continuously decrease until the end of training, whereas the validation loss  should similarly decrease for the first 50-100 epochs and then eventually plateau. Additionally, \texttt{tensorboard} will display the intermediate probability and distance predictions, which should converge to reasonable values. 
\item[] \textbf{? TROUBLESHOOTING}
\item Use the trained model to predict on new (full) raw stacks using the prediction code within the notebooks at \url{https://github.com/betaseg/protocol-notebooks/blob/main/stardist/run_stardist.ipynb}. Depending on the size of the stack, the prediction step may take 0.2-1h.
  
\end{steps}

\subsubsection*{Procedure 2e - Manual Skeleton Tracing - TIMING: 40-60 hrs} 
\figKnossos
\begin{steps}
\item Open \knossos and choose the dataset you want to work with~(\cref{fig:knossos}). Press OK.
\item Make sure that you have selected "6. Tracing Advanced" as work mode. Also open the cheat sheet by clicking on Help \menu Cheatsheet.
\item Navigate through the dataset until you find a microtubule. Move to one of the ends of the microtubule and start to set nodes by right mouse click.
\item After setting the second node \knossos will connect the nodes. After reaching the end of the microtubule press the C key to make a new tree (microtubule) and continue setting the nodes for the next microtubule.
\item Your annotation is automatically saved as XML file, which is later used by Album for the spatial analysis.%
\end{steps}
\subsubsection*{Procedure 2f - Curation of Segmentation Results - TIMING: 4-12 hrs}

No segmentation result will be absolutely perfect and we generally advise to proof-read each segmentation mask ideally by an expert different from the original annotator.
For mask curation, we recommend using either the Labkit plugin in \Fiji or ilastik (but your preferred image analysis software might suffice).\\

\emph{For mask curation in Labkit: }
\begin{steps}
\item Open \Fiji and load your dataset.
\item For making annotations in Labkit it is recommended to change the z-dimension of your dataset to time [t]. You can do this by clicking on Image \menu Properties and set Slices (z) to 1 and exchange the Frames (t) value with the number of z-slices. Press OK. Do the same also for the segmentation mask.
\item To load the raw data in Labkit go to Plugins \menu Labkit \menu Open Current Image With Labkit.
\item After the Labkit window has opened and you see the raw data you can load the corresponding segmentation mask by clicking Labeling \menu Open Labeling. Choose the file and click OK.
\item You can now navigate through the data with the left/right arrow keys and make corrections where needed. You can change the brush size and use the "Draw" and "Erase" tool.
\item Save your changes by clicking on Labeling \menu Save Labeling. Make sure to save the file as TIFF.
\end{steps}

\emph{For mask curation in ilastik:}
\begin{steps}
\item Open ilastik and create a new pixel classification project as shown before.
\item Open the raw data and proceed until step "3. Training".
\item Go to Advanced \menu Labels \menu Import Labels... and choose the corresponding segmentation file. Press OK.
\item You can now alter the labeling file with the "Brush Cursor" and "Eraser Cursor" tools. You should also give a few annotations for the background.
\item Click on "Live Update" and run the prediction.
\item You can then export only the corrected label file in "4. Prediction Export" by choosing Labels as source.
\item After clicking export you will save your desired segmentation file and the few labels for the background
\end{steps}

\subsection*{Procedure 3 - Volume and Spatial Analysis}

In our previous work the analysis of spatial interactions between organelles were performed manually. Here, all steps are run within the Album workflow, which simplifies the installation and execution of steps spanning across multiple tools and languages. The most recent documentation of this workflow can be found on \url{https://github.com/betaseg/cellsketch}. This includes command line calls for each step. The following parts of the protocol show the full analysis workflow performed via the graphical user interface (GUI) of Album.

\begin{demodata}{Spatial analysis for beta cell organelles}

    If you want to try out Procedure 3 without gathering your own data first, we provide example demo data for spatial analysis of beta cell organelles.
  In particular, we here use the dataset collection \lstinline{high_c1} from one of the cells this protocol was originally designed. You can access this dataset at \url{https://desycloud.desy.de/index.php/s/oxrzSiKbomiA9CZ}.
  This dataset has an isotropic pixel size of 0.016~\um.

  \begin{center}
    \includegraphics[width=.8\textwidth]{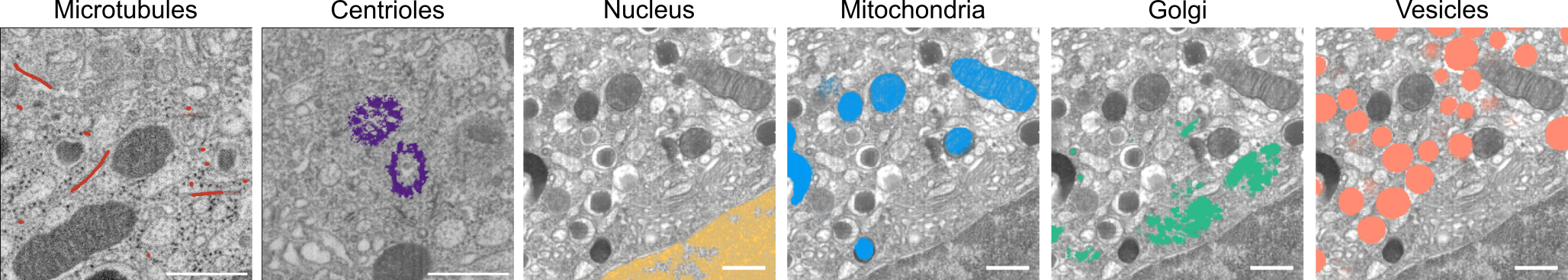}
  \end{center}
  {\small Panel adapted from [10] under CC BY 4.0 license.}
  
\end{demodata}

\figanalysisimport

\subsubsection*{Procedure 3.1 - Visualizing cell components with \cellsketch - TIMING: 20 min}

\begin{steps}
\item In order to be able to follow the full analysis and segmentation protocol, please make sure you have these datasets available from procedure 2~(\cref{fig:analysisimport}):
\begin{itemize}
\item Masks (\eg TIFF, pixel value zero for background, pixel value 1 or 255 for foreground). According to procedure 2 we expect the membrane mask to be filled. Furthermore the pixel size has to be isotropic, \eg 0.016~\um for the demo data. 
\item Labelmaps (\eg TIFF, each object has the unique pixel value, 0 for background) of the vesicles. As before, the pixel size has to be isotropic.
\item The labeled microtubules exported from \knossos as XML. As the \knossos analysis might have used raw images of higher resolution, ensure that the scale factors of the XML correspond to the lower pixelsizes of the masks. \Eg for the demo data the original microtubule segmentation was done on images with a pixel size of \sizethree{4~nm}{4~nm}{3.4~nm} such that the scale factors need to be \lstinline{[X=0.25, Y=0.25, Z=0.2125}] to result in the desired  target isotropic pixel size of 0.016~\um.
\end{itemize}
\item Install Album using the install wizard (instructions here: \url{https://docs.album.solutions/en/latest/installation-instructions.html}), this generates a launcher for Album either on your desktop or in the list of available applications. After launching Album, click on \menu catalogs in the bottom left corner, then click \menu Add catalog. Enter the following URL: \url{https://gitlab.com/album-app/catalogs/helmholtz-imaging} and confirm.
This is necessary for installing and running the specific solutions mentioned in the following steps. When running a solution which is not installed yet, you will be requested to install it first. You need to confirm an installation whenever one of the solutions is launched for the first time.

In order to visualize all segmentations and analysis values of individual organelle and microtubule instances in the same view, the data is transformed into a \cellsketch project. \cellsketch is a collection of solutions for displaying and analyzing single cell components.

\item After launching Album use the search bar or scroll through your list of solutions to find and run the \textbf{CellSketch: Create new project} solution. You then need to provide the following parameters:

\begin{itemize}
  \item \lstinline{parent}: Choose a folder where the project is going to be created in.
  \item \lstinline{name}: This name will represent the project, for the demo data you can call it \lstinline{high_c1}.
  \item \lstinline{input}: One dataset needs to be provided as the source image, for example your raw dataset from the data acquisition. It will only be used for visualization purposes as the background image. If none exists, one can simply choose one of the available masks. Any image file format that ImageJ / Fiji can open is expected to work.
  \item \lstinline{pixel_to_um}: The conversion factor from pixel units to µm units.
\end{itemize}

Since spatial analysis and visualization are performed, the imported datasets need to be scaled to be isotropic. Use the \lstinline{scale_z} parameter to provide a scaling factor for the z-axis if needed.

\item A new folder in the specified \lstinline{parent} folder with the ending \lstinline{.n5} will be created. Please do not rename it.

\item Next, the project will be displayed in the \cellsketch viewer. 

\item The viewer will initially only display the raw dataset. The view supports arbitrary rotations in 3D. Hover over the right part of the display - an arrow on the right border will appear. Click on it to show the sidebar of the viewer. After importing other cell components, they can be loaded and displayed by clicking on the eye symbols.

\item On the top of the right sidebar, there is a button called \menu Add dataset. It provides the following options:
\begin{itemize}
    \item \textbf{Add mask}: Mask datasets have the value 0 as background and 1 or 255 as foreground. They mark a component of the cell, like the nucleus, without distinguishing between multiple entities of the same component.   
    \item \textbf{Add labels}: Labels can be imported as label masks. Multiple entities of the same component type can be encoded by giving each object a unique pixel value whereas 0 is used for marking the background. A label mask does not support overlapping labels.
    \item \textbf{Add boundary}: The boundary of the cell is a mask but plays a special role when analyzing the data. It describes the space which is available for components within the cell. Therefore, the boundary needs to be a filled mask, not just the membrane itself. \cellsketch will automatically compute the outside border of this mask, add it as a cell component and call it `membrane`.
    \item \textbf{Add filaments from \knossos}: In case you used \knossos to annotate filaments, they can be imported using this option. The filaments are processed in the following fashion before being added to the project:
    \begin{itemize}
        \item Since in our experience \knossos ignored the first z-slices of the dataset without annotations, we add this offset based on comparing the number of z-slices of the \knossos file with the number of z-slices of the source dataset of the project.
        \item \knossos annotations contain line elements bundled as "things". Each "thing" is initially considered a filament. The line elements are not necessarily in the right order which makes it difficult to compute analysis on the filament ends. Therefore, we sort the line segment of each "thing" and split them into multiple filaments in case there are more than two line ends without another line end of the same group close by. The resulting list of points per filament is stored separately in YAML format and is the basis of further analysis.
        \item In order to compute distances and render the microtubules, \cellsketch stores a labelmap based on the computed point list per filament stored as YAML as previously described. The labelmap can be displayed in the \cellsketch viewer and later rendered just as the other labelmaps.
    \end{itemize}
\end{itemize}

All options share the following parameters:
\begin{itemize}
    \item \textbf{Name}: The name of the cell component. Will be used for displaying the component, in table columns and in file names i.e. when exporting meshes.
    \item \textbf{Scale factor X/Y/Z for dataset}: In case the dataset needs to be scaled to match the (isotropic) dimensions of the source datasets, these scaling parameters can be used.
    \item \textbf{Color}: The dataset will be displayed using this color - it can be later adjusted in the viewer by clicking the colored rectangle on the right side of the dataset name in the list of components in the right sidepanel. 
\end{itemize}
For adding labels, masks and the boundary, the following additional parameters are provided:
\begin{itemize}
    \item \textbf{Analyze connection to filaments ends}: Check this box if your project contains filaments and if you want the connection of their ends to the labelmap or mask to be analyzed.
    \item \textbf{Threshold to count filament ends as connected in µm}: When calculating the distance between a mask or label with filaments, this threshold in micrometers is used to mark filaments as connected to this mask/label.  
\end{itemize}
 
\textbullet \textbf{CRITICAL STEP} Adjust the microtubules scale values according to the correct voxel dimensions and binning factor of the raw data.\\

\item To delete an imported dataset, click on the three dots next to the bold name of the component in the sidepanel on the right and click \menu Delete.

\item In order to display the project at a different time, use the search bar or scroll to the solution called \textbf{CellSketch: Display data in BigDataViewer} in the Album interface and run it. Provide the newly created \lstinline{.n5} directory from the previous step as the \lstinline{project} input parameter.

\end{steps}

\figcellviewer

\subsubsection*{Procedure 3.2 - Spatial analysis - TIMING: 30 min}
  
Spatial analysis can be performed by directly clicking the \menu Analyze button in the right sidepanel of the viewer, however we highly advise running the analysis from a separate solution without the viewer being opened. The analysis can be very memory consuming.

\begin{steps}
  
\item Run spatial analysis for your \cellsketch project via the GUI by using the search bar or scrolling to the solution called \textbf{CellSketch: Run spatial analysis}. Run the solution with the following parameters:
\begin{itemize}
    \item \lstinline{project}: Your \cellsketch project, the directory ending with \lstinline{.n5}.
    \item \lstinline{connected_threshold_in_um}: When analyzing how close two organelles need to be in order to be counted as connected, this is the threshold, provided in micrometers.
\end{itemize}
\item[] \textbf{? TROUBLESHOOTING}\\

\item All results of the analysis are stored into \lstinline{MY_PROJECT.n5/analysis}. It will perform the following steps:
\begin{itemize}
  \item Distance maps for all imported components are computed. In a distance map, all pixel values represent the shortest distance of this pixel to a label or mask foreground.
  \item The mean, standard deviation (stdev) and median size of the labels of all labelmaps are computed and stored in \lstinline{PROJECT_NAME_LABELMAP_NAME.csv}.
  \item The distance and connectivity of all labels of all labelmaps to all masks and other labelmaps are computed and stored in \lstinline{PROJECT_NAME_LABELMAP_NAME_individual.csv} individually for each label. The number of connected vs. the number of not connected labels are stored in \lstinline{PROJECT_NAME_LABELMAP_NAME.csv}. This step includes the labelmaps of filaments - all filament pixels are considered, not just the filament ends.
  \item If filaments are present, the mean, stdev and median length and tortuosity of the filaments are stored in \lstinline{PROJECT_NAME_FILAMENTS_NAME.csv}. 
  \item If filaments are present, based on the parameters in the previous step, the distance between their ends and other labels / masks is computed and stored in\
  
  \lstinline{PROJECT_NAME_FILAMENTS_NAME_individual.csv} individually for each filament, the number of connected vs not connected filaments are stored in\
  
  \lstinline{PROJECT_NAME_FILAMENTS_NAME.csv}.
\end{itemize}
\end{steps}
\subsubsection*{Procedure 3.3 - Visualizing spatial analysis in the \cellsketch viewer -  TIMING: 10 min}

\begin{steps}
\item Launch the \cellsketch viewer solution (Album \menu \textbf{CellSketch: Display data in BigDataViewer}) to display the spatial analysis results.

\item Hover over the right part of the display - an arrow on the right border will appear - click on it to show the sidebar of the viewer. 

\item In the displayed list of all cell components, click on the eye symbols of the specific analysis component which should be visualized~(\cref{fig:cellviewer}). For example, displaying a distance map (i.e. the Golgi distance map) will show pixels further away from the respective target (in this case the Golgi) in a more opaque white color. Displaying the attribute of a specific label, for example the distance of SGs to the membrane, will color each label in respect to the value of the attribute for this specific object. All attributes of a specific label are also displayed as text in the top right corner of the viewer after clicking on it.

\end{steps}

\begin{demodata}{\cellsketch project for beta cell organelles including analysis}
    If you want to experiment with the plotting and visualization components of this protocol, you can download the \cellsketch project for the dataset collection \lstinline{high_c1} from one of the cells this protocol was originally designed for which already includes all analysis results. You can access this dataset at \url{https://www.dropbox.com/s/vs6bhkrozui7uob/mycell.n5.zip?dl=0}.  
\end{demodata}

\subsubsection*{Procedure 3.4 - Exporting connectivity masks - TIMING: 10 min}
The \cellsketch viewer can be used as a tool to generate TIFF masks of labels connected to specific masks. This can be useful for further analysis or visualization purposes.
\begin{steps}
\item Launch the \cellsketch viewer solution (Album \menu \textbf{CellSketch: Display data in BigDataViewer}) to display the spatial analysis results.
\item Hover over the right part of the display - an arrow on the right border will appear - click on it to show the sidebar of the viewer. 
\item In the displayed list of all cell components, scroll to the entry of a labelmap dataset, for example "microtubules". There should be connectivity entries listed below, for example "connected to centrioles" in case you added a centrioles mask. 
\item Click on the three dots on the right side of a connectivity entry, and click \menu Export mask. Click \menu Export inverted mask to export the labels which are not connected. 
\item The masks are exported as TIFF. The progressbar in the ImageJ main panel which opened alongside the \cellsketch viewer displays the export status. Once it is finished, the masks can be found in \lstinline{MY_PROJECT.n5/export}.
\end{steps}

\subsubsection*{Procedure 3.5 - Generating plots - TIMING: 15 min}

We provide a Jupyter notebook (\url{https://github.com/betaseg/protocol-notebooks/blob/main/plots/run_plots.ipynb}) showing how plots can be generated based on the analysis from the previous step. When running the notebook on the provided demo data, it should run without adjustments. 

You can clone the notebook with git and run it yourself, or you can use Album to run the notebook automatically in the correct environment. 

\begin{steps}
    \item After launching Album, use the search bar or scroll to the solution called \textbf{CellSketch: Plot analysis results}. Run the solution with the following parameters:
\begin{itemize}
    \item \lstinline{project}: Your \cellsketch project, the directory ending with \lstinline{.n5}.
    \item \lstinline{output}: The directory where plots generated in the notebook will be saved to~(\cref{fig:analysisplots}). This will also contain a copy of the notebook which can be further adjusted to generate more plots or to adjust the plots to a specific project.
\end{itemize}
    
    \item When running this command for the first time, the provided exemplary notebook will be executed and a copy of the notebook, including the result of the execution, will be stored in the provided \lstinline{output} directory. Afterwards and when running the solution with an output directory which already contains the notebook, the solution will initiate a new Jupyter notebook session and open the browser with the output Jupyter interface where one can click on \lstinline{plots.ipynb} to adjust the notebook.
\end{steps}

\textbullet \textbf{CRITICAL STEP} If the project has different cell component names than the exemplary dataset, exceptions will occur. Please adjust the code in the notebook to generate plots based on the naming of the cell components in your project.

\figanalysisplots

\subsection*{Procedure 4 - Visualization}

In the following paragraph we provide instructions on visualization of vEM segmentation data. We specifically describe how to generate overlays of raw and segmentation data with 3Dscript and how to produce 3D renderings with ORS Dragonfly and Blender. The steps for generating renderings in Blender are provided as Album solutions and integrated into the \cellsketch workflow. 

\begin{demodata}{Visualization of beta cell organelles}

    If you want to try out Procedure 4 without gathering your own data first, we provide example demo data for spatial analysis of beta cell organelles.
  In particular, we here use the dataset collection \lstinline{high_c1} from one of the cells this protocol was originally designed. You can access this dataset at \url{https://desycloud.desy.de/index.php/s/oxrzSiKbomiA9CZ}.

  \begin{center}
    \includegraphics[width=1\textwidth]{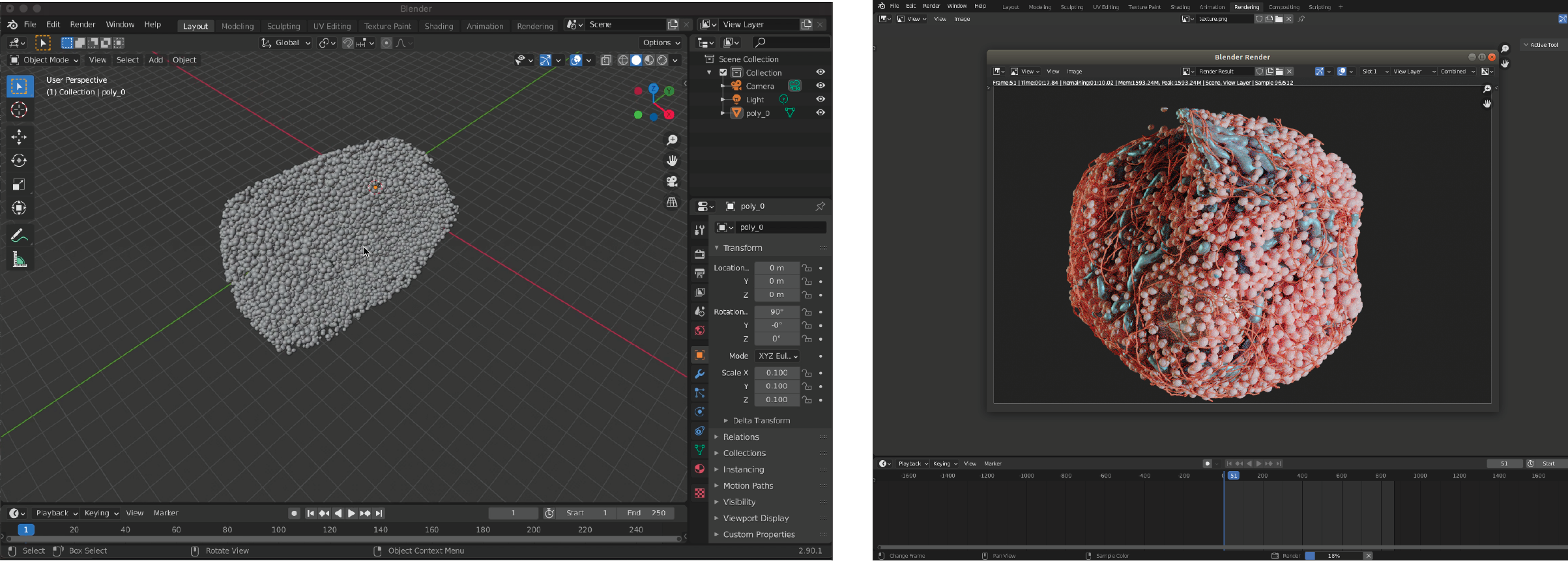}
  \end{center}
  
\end{demodata}

\subsubsection*{Procedure 4a - Overlay visualization with 3Dscript - TIMING: 15-30 min}
Within the protocol we used 3Dscript mainly for visualizing the two dimensional overlays of raw and segmentation data (\cf Demo Data 3)\footnote{Note, however, that 3Dscript can also be used to create 3D renderings and animations}. To make the overlays follow these steps:

\begin{steps}
\item Open the raw files and the segmentation files you want to visualize in \Fiji.
\item Go to Image \menu Color \menu Merge Channels.
\item Define the colors for each channel (these do not have to be your final colors, you can set them later in 3Dscript) and click OK.
\item Go to Plugins \menu 3D script \menu Interactive Animation.
\item \Fiji will now open the 3Dscript window and the "Interactive raycaster" and you will see a 3D rendering of your file.
\item Change the Rendering algorithm to "Combined transparency".
\item Now change the color of each organelle to your pre-defined one by double-click on the color block next to the channel and double-click on the colored rectangle that appears. You can now set the HTML color code for the "Foreground color".
\item To make 2D overlays of raw and segmentation data go to Cropping \menu Show and set the z-range to only a small value of 1-2 slice. Press Enter and a 2D view will appear. Zoom in if necessary.
\item Export an image of the area in the window by clicking Animation \menu Show \menu Start text-based animation editor.
\item Go to Record \menu Record transformation and set the frame to 1.
\item Click "Run" and save the result as TIFF or PNG file.
\item You can also make complex 3D animations with the text-based editor and by changing the values of the axes of the organelle channels. We recommend to watch the 3Dscript tutorial video~(\url{https://bene51.github.io/3Dscript/gallery.html}).
\end{steps}
  
\subsubsection*{Procedure 4b - Visualization with ORS Dragonfly -  TIMING: 60-90 min}

\begin{steps}
\item Open ORS Dragonfly.
\item Load your individual raw and segmentation masks by going to File \menu Import Image Files... \menu Add... and choose the file you want to open. Click "Next" and define the voxel dimensions of your dataset. Click Finish. Repeat this step for all the individual masks.
\item Right-click on the image and select 3D for rendering.
\item On the left panel you can change the "Background color" in the "Scene's View Properties" tab. You may change it to white or black.
\item Now set the colors for the organelles according to your color palette~(\cref{fig:3D}a). You do so by clicking on your file in the "Data Properties and Settings" tab. Go to the "Window Leveling" tab on the left side. Set one of the preset colored LUTs (\eg green). Now right-click on the colored "Value Box" and change the HTML value as it is defined in your palette. Click OK. Set the levers below the histogram in a way to achieve a uniform color. Now click "More..." next to LUT and define a name (ideally the organelle name) for the new histogram. Repeat this step for all organelles.
\item In order to make animations or export single frames~(\cref{fig:3D}b) right-click on the main window and click "Show Movie Maker". Now additional options below the main window will appear that allows the setting of key frames. For creating animations it makes sense to activate the "Clip box" by pressing the "Clip" button in the "Clip" tab. You can then rotate, move or clip the data and add new key frames by pressing "Add Key".
\item You can run the animation by pressing "Play" and once you satisfied with export it by clicking on the "Export Animation" button. Choose the desired speed, pixel dimensions and file format and press "Save as Video File".
\end{steps}

\figvtk

\subsubsection*{Procedure 4c - Visualization with VTK and Blender - TIMING: 1h}

\begin{demodata}{\cellsketch project for beta cell organelles}
    You can download an existing \cellsketch project to run this procedure based on the dataset collection \lstinline{high_c1} from one of the cells this protocol was originally designed for. You can access this dataset at \url{https://www.dropbox.com/s/vs6bhkrozui7uob/mycell.n5.zip?dl=0}.  
\end{demodata}

We provide Album routines to automatically generate meshes from segmentations and labels masks, import of the meshes into Blender, apply colored materials to each component and generate a scene which can be rendered in high quality. Before the full rendering in Blender results can be quickly evaluated in the visualization toolkit (VTK).
\begin{steps}
\item Install Album using the install wizard (instructions here: \url{docs.album.solutions/en/latest/installation-instructions.html}), this generates a launcher for Album either on your desktop or in the list of available applications. After launching Album, click on \menu catalogs in the bottom left corner, then click \menu Add catalog. Enter the following URL: \url{https://gitlab.com/album-app/catalogs/helmholtz-imaging} and confirm.
This is necessary for installing and running the specific solutions mentioned in the following steps as described before. 
\item After launching Album, use the search bar or scroll to the solution called \textbf{CellSketch: Export masks and labelmaps as meshes}. Run the solution with the following parameter:
\begin{itemize}
    \item \lstinline{project}: Your \cellsketch project, the directory ending with \lstinline{.n5}.
\end{itemize}

The optional parameters \lstinline{include} and \lstinline{exclude} provide ways of choosing a subset of datasets. Only components containing the \lstinline{include} string in their name will be exported, dataset names containing the \lstinline{exclude} string will be ignored. Multiple inclusion strings can be provided by separating them by a comma, for example \lstinline{mito,nucleus}. 

\item The meshes generated in the previous step were saved to \lstinline{MY_PROJECT.n5/export/meshes} or the respective folder in your \cellsketch project directory. Either open Blender and import all generated STL files manually via \menu File \menu Import \menu Stl or use the following solution to import all meshes jointly from a folder. The script will also adjust the viewport of the camera to ensure all meshes are visible and assign a glass material in the color as specified in the \cellsketch project to each cell component. 

Import your meshes into Blender via GUI by using the search bar or scrolling to the solution called \textbf{CellSketch: Generate Blender scene from mesh files}. Run the solution with the following parameters:
\begin{itemize}
    \item \lstinline{project}: Your \cellsketch project, the directory ending with \lstinline{.n5}.
    \item \lstinline{output_blend}: The path to where the Blender project file will be stored, useful for further adjustments of the scene.
\end{itemize}

\figrendering

You can manually assign or edit materials to each object by clicking on the \menu Shading workspace in the top center area, selecting the object on the right and adjusting or adding a material in the lower part of the user interface.

\item You can now open the Blender project stored at \lstinline{output_blend} with Blender - either you already have Blender installed, then just open it there, or use the Album solution by launching Album, using the search bar or scrolling to the solution called \textbf{Launch Blender 2.83.13}.

\item Click \menu Render \menu Render image to render the cell~(\cref{fig:3D}c). Adjustments to the rendering can be done after selecting the \menu Rendering workspace in the top center area. Select \lstinline{Cycles} in the right area as Render Engine for enhanced material rendering. GPU devices can be configured by clicking \menu Preferences \menu System \menu Cycles Render Devices.
\end{steps}

\subsubsection*{Procedure 4d - Check colors for color blind friendliness}
In all of the procedures 4a-c above one can define a color palette for the organelles, which should be used throughout the project~(\cref{fig:3D}a). It is important that this color palette is adjusted to be colorblind friendly. 
You can check the palette file or one of the 3D renderings for colorblind friendliness by
\begin{steps}
\item Opening the file in \Fiji.
\item Go to Image \menu Color \menu Simulate Color Blindness.
\item Define the type of colorblindness you want to check and press OK.
\item You will see how your colors will appear to people with the specific kind of colorblindness.
\item Change your palette if the colors cannot be easily discriminated.
\end{steps}

\section*{Troubleshooting}

For troubleshooting see~\cref{table:1}.

\begin{table*}[h]
\centering
\footnotesize
 \def\arraystretch{.8}
\begin{tabular}{ p{1.8cm} p{1.3cm} p{3.6cm}  p{3.6cm} p{4.2cm}} 
 
 \toprule
 \textbf{Procedure} & \textbf{Step} & \textbf{Problem} & \textbf{Possible reasons} & \textbf{Solution}\\
  \midrule
 1a & 1 & Full vEM volume cannot be opened. & Raw data are too large for computer memory. & Open file as virtual stack in \Fiji or IMOD, or bin file before preparing crops. \\ 
  \midrule
 2a + 2b  & 11 + 14 & Imported Segmentation mask is black in \Fiji. & You forgot to perform Object Detection in ilastik or you did not yet convert the file into a binary mask in \Fiji. & Perform the Object Detection step in ilastik by clicking once on the desired organelle and press "Live Update". After importing the mask into \Fiji go to Process \menu Binary \menu Convert to Mask and choose a method for binarization. Now your labels should be visible. \\
  \midrule  
 2b & 7 & Ilastik probability map does not look reasonable. & Not enough labels for training. & Increase the number of labels or add more classes in the first Autocontext classification step. \\
  \midrule 
 2c & 14 & Validation loss is heavily increasing after a certain number of epochs (overfitting)  & Indicates that training data is not sufficient or was wrongly annotated & Check your ground truth and correct it. Prepare more crops from diverse regions of the cell (also with negative example and label them precisely. Add more data augmentation.\\ 
  \midrule 
 2d & 6 &  &  same as above &  \\ 
  \midrule 
 3.2 & 1 & Spatial analysis crashes or unexpectedly quits before finishing. & The memory of the system is not sufficient to perform the analysis. & Distance maps are most memory-expensive to compute. In case the analysis process crashes because of memory issues, repeat the analysis process and select the "skip existing distance maps" checkbox. It will not recompute already computed distance maps. This has to be unchecked whenever an existing dataset is deleted or changed. Otherwise, use a system with more RAM. \\
  \bottomrule
\end{tabular}
\caption{\textbf{Troubleshooting}}
\label{table:1}
\end{table*}

\section*{Timing}
The information on timing in procedure 1 is referring to one cell out of a large vEM stack with voxel dimensions of approximately \sizethree{7,500}{7,500}{5,000} pixels. For procedure 2 the timing is estimated for segmentation of one cell (approximately 1/5 of the original stack) binned with factor 4 (except for microtubule segmentation which was performed at the original resolution). Timing of processes 3 and 4 also refer to one cell of the original volume.\\
Procedure 1 - Preparation of Raw Data: 1a - 10-30 min, 1b - 5 min, 1c - 10-30 min, 1d - 10 min\\
Procedure 2 - Segmentation: 2a - 45-120 min, 2b - 4-24 hrs, 2c - 12-36 hrs, 2d - 12 -36 hrs, 2e - 40-60 hrs, 2f - 4-12 hrs\\
Procedure 3 - Spatial Analysis: 3.1 - 20 min, 3.2 - 30 min, 3.3 - 10 min, 3.4  -10 min, 3.5 - 15 min\\
Procedure 4 - Visualisation: 4a - 15-30 min, 4b - 60-90 min, 4c - 1h

\section*{Anticipated results}

This protocol provides modular steps for achieving a precise 3D segmentation of whole cells including various organelle classes out of vEM datasets followed by spatial analysis and 3D visualization.

At the beginning of the protocol we give advise on the preparation of the raw vEM data including binning and file conversion. After these first steps the data will have the optimal size and file format(s) for the following segmentation tasks.

The user should be able to decide which segmentation strategy matches best with their organelle of interest and is then able to perform these task in a time-efficient manner.

After successful segmentation and curation of the segmentation masks the user will be able to perform volume analysis of the organelles as well as investigate spatial interactions between different organelle classes.

Finally, 3D rendering of the segmentation results can be applied to generate publication-ready figures and videos.\\
The steps for spatial analysis, plotting of the results and 3D rendering with Blender are implemented in solutions which can be run across platforms by using album. This allows for achieving publication-ready plots and renderings within a few hours after finalizing the segmentation tasks.\\

The user can expect to get fundamental insights into their vEM data with information on the volume fractions, shapes and distribution of organelles within cells. When comparing different cell culture or disease conditions the analysis allows for addressing structural changes within cells in an unprecedented manner.
In our project~\cite{muller2021} we were able to evaluate heterogeneities between cells regardless of the culture conditions by calculating volume fractions of organelles. We could find differences in microtubule lengths between metabolic stimuli - a key hint that metabolism influences ultrastructure.
Spatial analysis enabled us to unravel the connectivity of microtubules in beta cells. They were largely non-centrosomal and surprisingly also not Golgi-connected. Measuring the 3D association of insulin SGs with microtubules allowed us to categorize SGs in "associated" and "not associated" and the finding that microtubule-SG interaction preferably occurs in the cell periphery. Plotting spatial data against random distributions further allowed us to investigate possible regulated processes in the cell, which we could observe in the case of the enrichment of insulin SGs and microtubules close to the plasma membrane. In addition to the plotted data, 3D renderings helped to experience the data more intuitively and to appreciate better the crowding of the cytoplasm, the different cell and organelle shapes as well as the basic organization of the microtubule scaffolds. In summary, our protocol provides a complete segmentation, analysis and visualization pipeline to fully embrace the complexity of modern vEM data.

\section*{Reporting Summary}
Further information on research design is available in the Nature Research Reporting Summary linked to this article.

\section*{Data availability}
All raw datasets at their original resolution are available at OpenOrganelle~( \url{https://openorganelle.janelia.org/}). Demo data of segmentation masks and for deep learning are available at \url{https://desycloud.desy.de/index.php/s/oxrzSiKbomiA9CZ}.
Demo data for spatial analysis are available at \url{https://www.dropbox.com/s/vs6bhkrozui7uob/mycell.n5.zip?dl=0}.
\section*{Code availability}
The code for Album is available at \url{https://gitlab.com/album-app/album}. 
The code for the Album solutions is available at \url{gitlab.com/album-app/catalogs/helmholtz-imaging}. 
The code for the \cellsketch viewer is available at \url{github.com/betaseg/cellsketch}.
The jupyter notebooks for demo workflows can be found at \url{https://github.com/betaseg/protocol-notebooks}.

\begin{acknowledgements}
We thank Katja Pfriem for administrative assistance. We thank members of the PLID for valuable feedback. We thank Song Pang and C. Shan Xu (Yale University, USA) as well as Harald F. Hess (Janelia Research Campus, USA) for FIB-SEM. We thank Susanne Kretschmar and Thomas Kurth (CMCB, Germany) for initial sample preparation. We thank all further authors of the original JCB publication, Joyson Verner D'Costa, Carla Münster (both PLID, Germany), and Florian Jug (Human Technopole, Italy) for their support. This work was supported by the Electron Microscopy and Histology Facility, a Core Facility of  the CMCB Technology Platform at TU Dresden. We thank the EM facility of the Max Planck Institute of Molecular Cell Biology and Genetics for their services. This work was supported with funds to MS from the German Center for Diabetes Research (DZD e.V.) by the German Ministry for Education and Research (BMBF) and from the Innovative Medicines Initiative 2 Joint Undertaking under grant agreement No 115881 (RHAPSODY). This Joint Undertaking receives support from the European Union’s Horizon 2020 research and innovation program and EFPIA. This work is further supported by the Swiss State Secretariat for Education‚ Research and Innovation (SERI) under contract number 16.0097-2. AM was the recipient of a MeDDrive grant from the Carl Gustav Carus Faculty of Medicine at TU Dresden.
MW was supported by the EPFL School of Life Sciences and a generous foundation represented by CARIGEST SA.
DS and LR were funded by HELMHOLTZ IMAGING, a platform of the Helmholtz Information \& Data Science Incubator.
\end{acknowledgements}

\begin{contributions}
A.M., D.S., M.S. and M.W. wrote the manuscript. D.S. and M.W. wrote the workflow implementations. L.R. tested the workflows and edited the manuscript. 
\end{contributions}

\begin{interests}
The authors declare no competing interests.
\end{interests}

\section*{Related Links}
\textbf{Key references using this protocol}\\
Müller, A. et al. \textit{J. Cell. Biol.} 220 (2021):~\url{https://doi.org/10.1083/jcb.202010039}
\newpage

\printbibliography

\newpage

\end{document}